\documentclass[lettersize,journal]{IEEEtran}

\usepackage[utf8]{inputenc} 
\usepackage[T1]{fontenc} 

\usepackage{amsfonts} 
\usepackage{amsmath} 
\usepackage{bm} 
\usepackage{mathtools} 
\usepackage{nicefrac} 

\usepackage{graphicx} 
\usepackage{subfig} 

\usepackage{array} 
\usepackage{booktabs} 
\usepackage{colortbl} 
\usepackage{multirow} 

\usepackage{algorithm} 
\usepackage{algorithmic} 

\usepackage[nocompress]{cite} 
\usepackage{hyperref} 
\hypersetup{
 colorlinks = true,
 citecolor = blue,
 filecolor = magenta,
 linkcolor = blue,
 urlcolor = black
}
\usepackage[capitalise]{cleveref} 
\usepackage[numbers]{natbib} 

\usepackage{comment} 
\usepackage{microtype} 
\usepackage{textcomp} 
\usepackage{verbatim} 

\usepackage{xcolor} 
\usepackage{orcidlink} 

\usepackage{balance} 
\usepackage{stfloats} 

\def\BibTeX{{\rm B\kern-.05em{\sc i\kern-.025em b}\kern-.08em
 T\kern-.1667em\lower.7ex\hbox{E}\kern-.125emX}}

\begin{document}

\title{QIRL: Optimized Question-Image Relation Learning for Bias-Robust Visual Question Answering}

\author{
Quanxing~Xu\orcidlink{0009-0008-4354-8371}, 
Ling~Zhou\orcidlink{0000-0002-8313-5749}, 
Xian~Zhong\orcidlink{0000-0002-5242-0467},~\IEEEmembership{Senior~Member,~IEEE},
Feifei~Zhang\orcidlink{0000-0002-8153-9977}, 
Rubing~Huang\orcidlink{0000-0002-1769-6126},~\IEEEmembership{Senior~Member,~IEEE}. 

\thanks{Manuscript received April 4, 2025. This work is supported in part by the Science and Technology Development Fund of Macau, Macau SAR, ( Grants No. 0035/2023/ITP1 and 0021/2023/RIA1), the National Natural Science Foundation of China ( Grants No. 62271361 and 62376196), and the Hubei Provincial Key Research and Development Program ( Grant No. 2024BAB039), and the Tianjin Natural Science Foundation ( Grant No. 24JCJQJC00190). (\textit{Corresponding authors: Ling Zhou and Xian Zhong})}

\thanks{Quanxing Xu and Ling Zhou are with the School of Computer Science and Engineering, Macau University of Science and Technology, Taipa, Macau 999078, China (email: 3230002299@student.must.edu.mo; lzhou@must.edu.mo).}

\thanks{Xian Zhong is with the Hubei Key Laboratory of Transportation Internet of Things, School of Computer Science and Artificial Intelligence, Wuhan University of Technology, Wuhan 430070, China and State Key Laboratory of Maritime Technology and Safety, Wuhan University of Technology, Wuhan 430063, China (email: zhongx@whut.edu.cn).}

\thanks{Feifei Zhang is with the School of Computer Science and Engineering, Tianjin University of Technology, Tianjin 300382, China (email: feifeizhang@email.tjut.edu.cn).}

\thanks{Rubing Huang is with the School of Computer Science and Engineering, Macau University of Science and Technology, Taipa, Macau 999078, China (email: rbhuang@must.edu.mo).}


}

\markboth{IEEE TRANSACTIONS ON IMAGE PROCESSING, 2025}%
{How to Use the IEEEtran \LaTeX \ Templates}

\maketitle

\begin{abstract}

Existing bias mitigation methods for Visual Question Answering (VQA), a typical Artificial intelligence application, endure two main limitations. First, they fail to capture the optimal relation between images and texts, as prevailing learning frameworks lack the capacity to extract deep correlations from highly contrasting samples. Second, they overlook assessing Question-Image (QI) relevance during inference, since prior work has not examined the degree of input relevance in debiasing studies. To address these issues, we propose a novel neural network framework termed Optimized \underline{Q}uestion-\underline{I}mage \underline{R}elation \underline{L}earning (QIRL), which provides a reliable implementation of artificial intelligence for VQA tasks and improves the robustness of conventional VQA models through a generation-driven self-supervised learning strategy. Specifically, two modules are introduced. The \textit{Negative Image Generation} (NIG) module automatically produces highly irrelevant QI pairs during training to strengthen relational learning.
In contrast, the \textit{Irrelevant Sample Identification} (ISI) module enhances model robustness by detecting and filtering out irrelevant inputs, thereby reducing prediction errors. Moreover, to verify the effectiveness of filtering out unrelated QI pairs in mitigating output errors, we propose a specialized metric to evaluate the ISI module's performance. Notably, our approach is model-agnostic and can be seamlessly integrated with various VQA architectures. Extensive experiments on \textsc{VQA-CPv2} and \textsc{VQA-v2} datasets demonstrate the effectiveness and generalization ability of our method. Our approach achieves state-of-the-art performance. 

\end{abstract}


\section{Introduction}
\begin{figure}[!t]
	\centering
	\includegraphics[width = 0.9\linewidth]{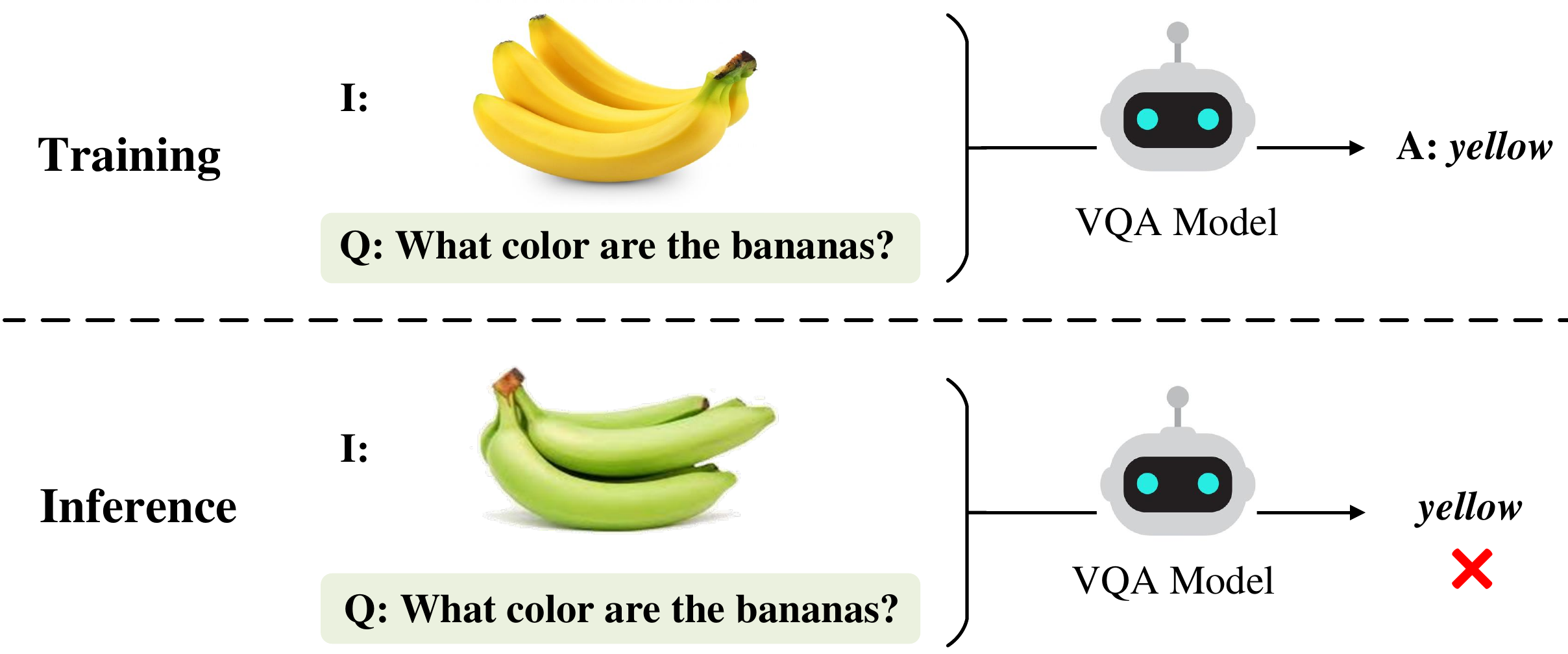}
	
	\caption{\textbf{Illustration of Language Bias.} VQA models often exploit superficial correlations between questions and answers, neglecting the image content during inference.}
	\label{fig1}
\end{figure}

\IEEEPARstart{V}{ision}-language tasks serve as representative benchmarks for evaluating models' capabilities in multi-modal learning and visual-linguistic understanding, including visual storytelling~\cite{zhongxian1}, video captioning~\cite{zhongxian2,zhongxian3}, and Visual Question Answering (VQA)~\cite{1}. As one of the most essential vision-language tasks, VQA, where an image is accompanied by a natural language question, has attracted considerable attention in recent years due to its requirement for generating accurate natural language responses. This growing interest reflects a shift in the image processing community from conventional "bucketed" recognition toward addressing more intricate multi-modal challenges~\cite{1,2}. Nevertheless, language bias remains a significant challenge in this field. For instance, as illustrated in \cref{fig1}, the dominant answer for the question type ``What color \dots bananas?'' in the training data is ``yellow''. Consequently, VQA models may exploit this simple correlation, relying on superficial cues rather than integrating visual information to capture underlying semantics and perform reasoning, which can lead to incorrect answers. 
Recently, although kinds of large models, such as large language models (LLMs), have been widely used in vision-language tasks (e.g., Knowledge-based VQA (KBVQA)~\cite{74,75}), language bias still exists within them and affects their reasoning and thereby weakening the reliability of the results. Latest studies~\cite{68,70,91} have demonstrated that the problem of prediction credibility remains in KBVQA methods employing LLMs, Pre-trained Language Models (PLMs), Vision-Language Pre-trained models (VLPs), and Multi-modal Large Language Models (MLLMs) due to the fact that biases (preferences) are inevitably present in the distribution of training data.

To address these issues for conventional VQA models, some studies focused on constructing more balanced training datasets to alleviate strong correlations between questions and answers (\textit{e.g.}, correlating ``Is there \dots ?'' with ``yes'', ``What color is \dots ?'' with ``black'', and ``How many \dots ?'' with ``2'')~\cite{1,2}. 
In addition, there are five kinds of debiasing methods: novel model architectures~\cite{11,13}, refined language model (LM)~\cite{53,55}, enhanced visual attention mechanisms~\cite{14,15}, ensemble methods~\cite{56,57}, and data-driven strategies~\cite{16,17}. 
Likewise, research on debiasing for VQA using large models started to emerge recently~\cite{68}.

\begin{figure}[!t]
	\centering
	\includegraphics[width = 0.9\linewidth]{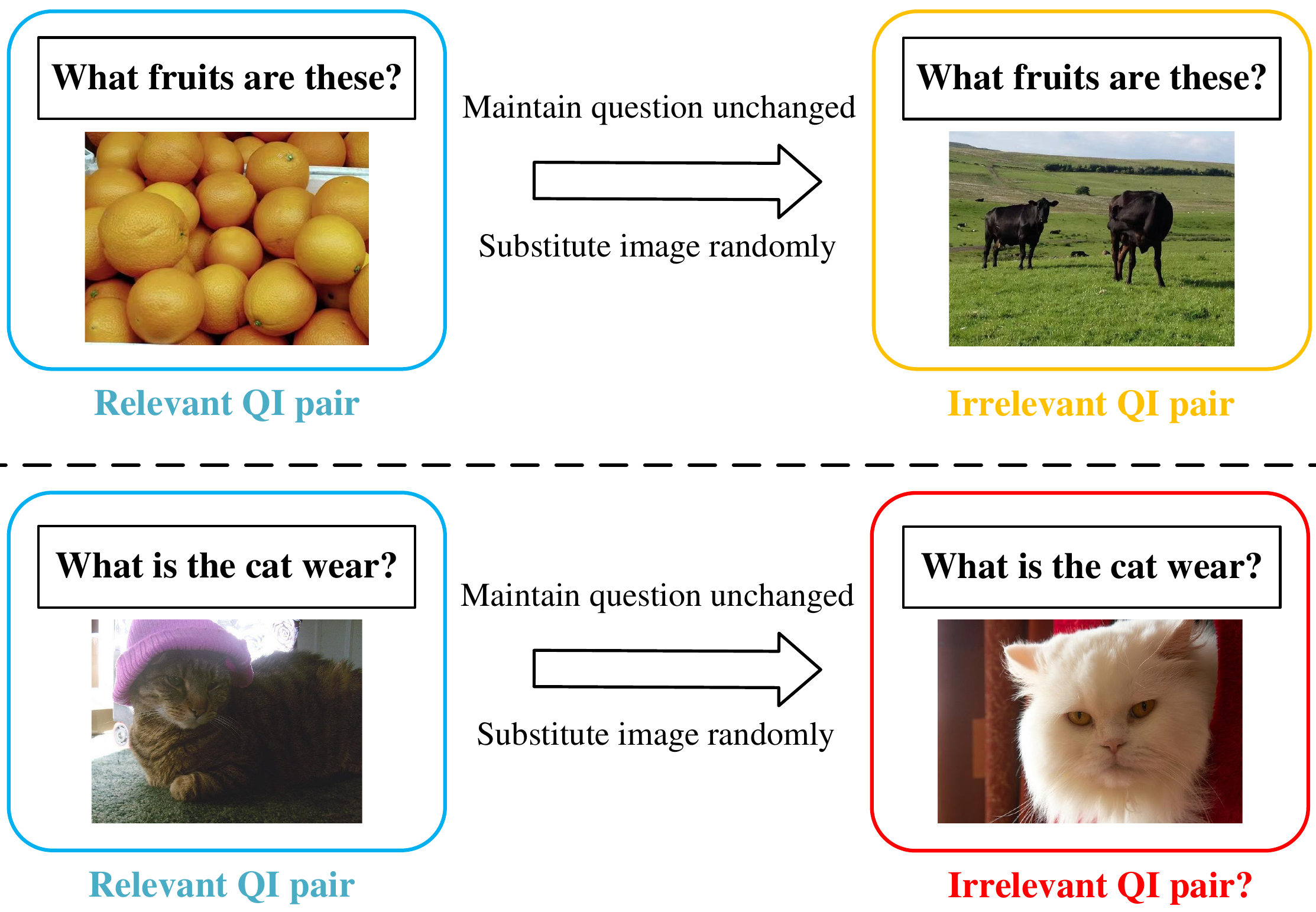}
	
	\caption{\textbf{Illustration of Existing Approach for Generating Irrelevant QI Pairs.} Although designed to produce irrelevant pairs, this method is flawed: as shown in the bottom row, both images contain a cat while the question pertains to a cat, rendering the generated pair relevant.}
	\label{fig2}
\end{figure}

In vision-language tasks, effective question-image (QI) relation learning is critical for high model performance, whereas irrelevant QI inputs can adversely affect predictions. To encourage models to learn authentic QI relations and avoid reliance on superficial correlations, existing methods generate irrelevant QI pairs by randomly substituting the image while keeping the question unchanged. In doing so, the model simultaneously learns the genuine QI correlation from the original pair and the superficial correlation from the generated pair, guided by an appropriate loss function. However, this approach may inadvertently produce QI pairs that remain relevant. As illustrated in \cref{fig2}, the bottom row shows both images containing a cat while the question pertains to the cat; thus, the generated QI pair is not truly irrelevant, which may adversely affect subsequent learning.

\begin{figure}[!t]
	\centering
	\includegraphics[width = \linewidth]{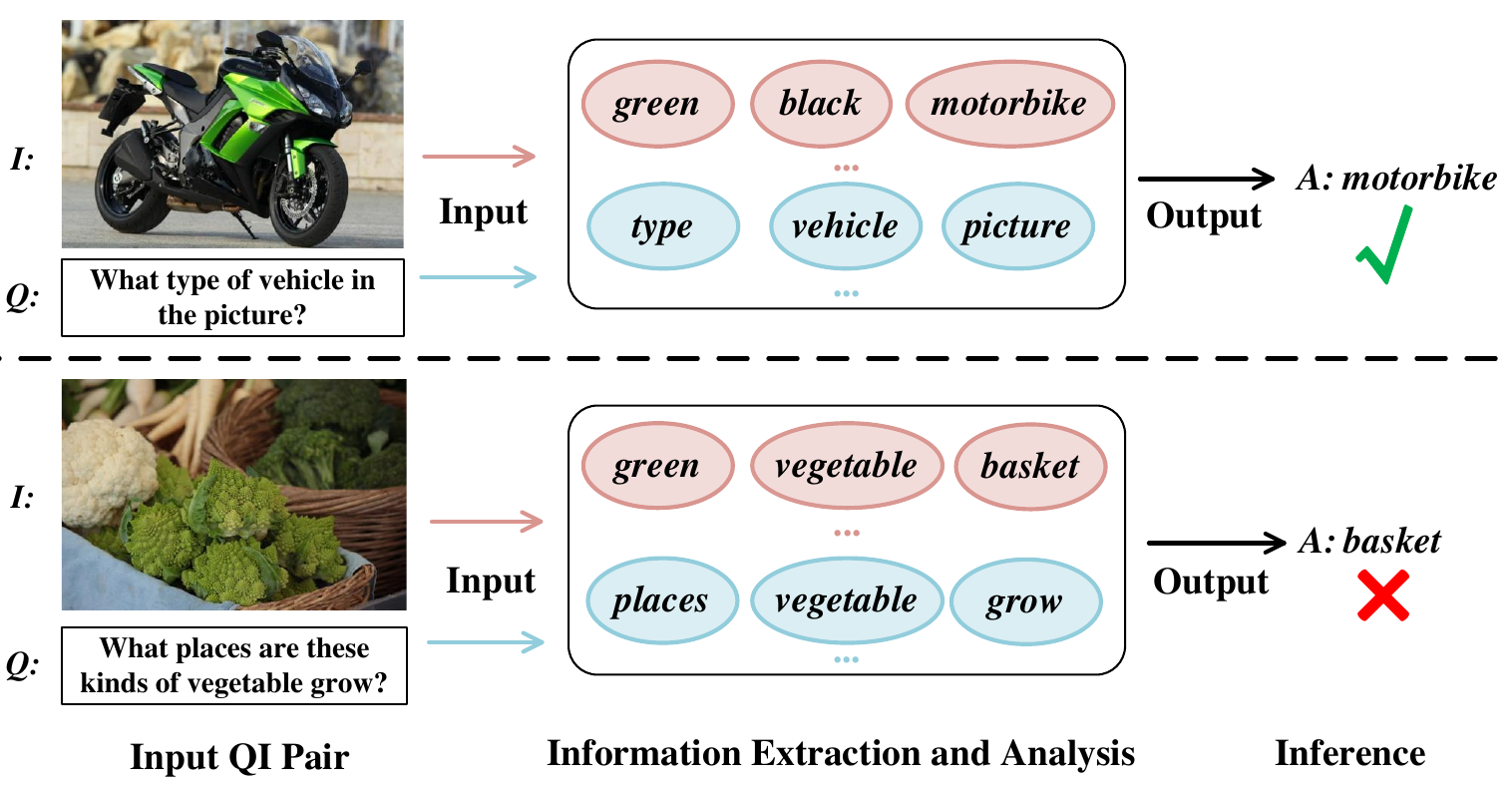}
	
	\caption{\textbf{Influence of Relevant and Irrelevant QI Pairs on Predictions.} The top instance represents a relevant QI pair, whereas the bottom instance represents an irrelevant QI pair. The red ellipse denotes visual information, and the blue ellipse denotes semantic information. Note that the information extraction and analysis depicted here is schematic and does not reflect the model's actual operation.}
	\label{fig3}
\end{figure}

For relevant inputs, the model can correctly infer the answer based on both the image content and the question's meaning. In contrast, when presented with an irrelevant QI pair, the model may be misled by mismatched information, leading to incorrect answers. As shown in \cref{fig3}, the relevant input (top instance) provides multi-modal information that facilitates correct reasoning, whereas the irrelevant input (bottom instance) offers unrelated cues that misguide the prediction. If the model refrains from answering irrelevant QI pairs, the overall error rate can be significantly reduced, thereby enhancing its robustness when encountering abnormal inputs. However, no debiasing study has yet addressed the identification of the relevance degree of QI inputs.

Motivated by these challenges, we posit that
1) refined QI relation learning can enhance model performance, and 
2) filtering irrelevant QI pairs before prediction can improve model robustness.
Accordingly, we propose Optimized \underline{Q}uestion-\underline{I}mage \underline{R}elation \underline{L}earning (QIRL), a generation-based self-supervised learning framework designed to strengthen QI relation learning via a fresh sample generation manner to produce more contrastive samples during the training and a novel identification module for filtering unrelated QI pairs before the model's inference.

Specifically, we introduce a Negative Image Generation (NIG) module that produces irrelevant QI pairs without human supervision, and an Irrelevant Sample Identification (ISI) module that enables VQA models to handle inputs that do not correspond to relevant QI pairs. The NIG module consists of two components: a diffusion model that converts the original image into a textual description, and an unsupervised sentence revision tool that transforms this description into a negative sentence. This negative sentence is then input to the diffusion model to generate a new image, yielding a QI pair that is markedly different from the original. 
The ISI module serves as a discriminator between common and irrational QI pairs, and is trained jointly with the VQA model using both original and generated QI pairs. After training, the module directs the model either to produce a prediction for a common QI pair or to output ``abstain'' when presented with an irrational QI pair.

Our key contributions can be summarized in threefold:

\begin{itemize}
	\item We design a novel generation approach for producing highly irrelevant QI pairs within a self-supervised framework for enhancing the model's QI learning from an absolute contrastive sample perspective.

	\item We propose an identification module to distinguish between relevant and irrelevant inputs in inference, significantly enhancing the model's robustness by isolating the adverse effect from unmatched inputs on accuracy.

	\item Extensive experiments on \textsc{VQA-CPv2} and \textsc{VQA-v2} datasets with various base VQA models demonstrate the notable generalizability of the proposed method. Besides, our method achieves performance comparable to state-of-the-art approaches.

\end{itemize} 

\begin{figure}[!t]
	\centering
	\includegraphics[width = 1.0\linewidth]{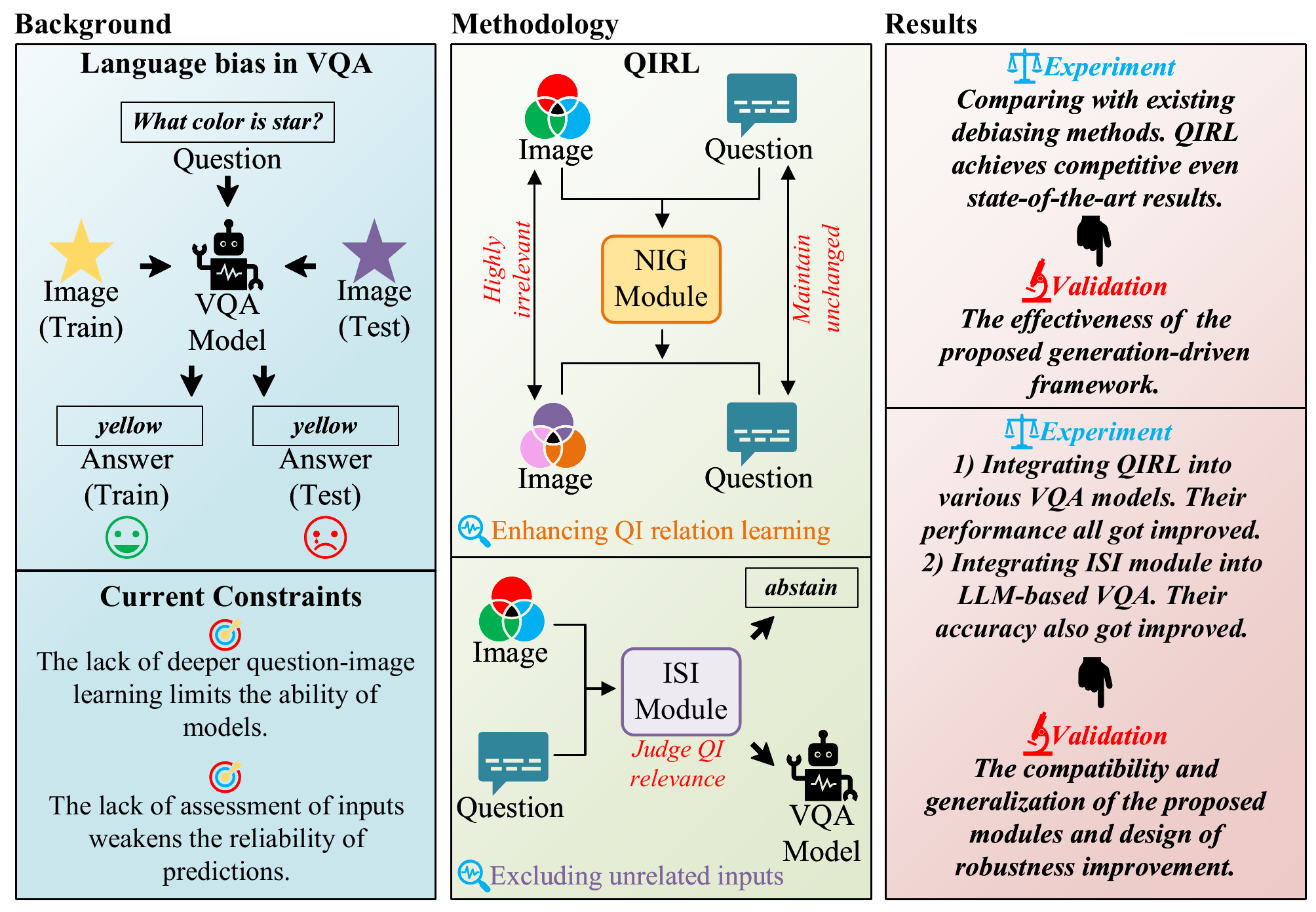}
	\caption{\textbf{Overall framework of the proposed QIRL method.}}
	\label{fig0}
\end{figure}

The remainder of this paper is organized as follows. Section \ref{sec:relate} reviews related work, Section \ref{sec:metho} details the proposed method, Section \ref{sec:experi} presents experiments, and Section \ref{sec:conclu} concludes and outlines future directions. The overall framework of our work is depicted in Section \ref{fig0}.

\section{Related Work}
\label{sec:relate}
\subsection{Visual Question Answering}

Given an image and a natural language question, Visual Question Answering (VQA)~\cite{1} requires a system to provide an accurate natural language response. VQA is widely recognized as a benchmark for evaluating models' capabilities in multi-modal learning and visual-linguistic understanding. In recent years, with the advancement of LLMs, Knowledge-Based VQA (KBVQA)~\cite{76,79,80,92} has emerged as a popular task that leverages the extensive knowledge of these models to address more challenging questions. Although large models demonstrate notable performance in VQA, they still suffer from language bias inherited from their training data. 
In this work, we aim to design a compatible debiasing framework for conventional VQA models, which alleviates language bias by optimizing question-image (QI) relation learning and introducing an innovative module to detect and filter irrelevant inputs, thereby enhancing model robustness.

\subsection{Debiasing Methods for VQA}
Language bias, arising when the answer distribution in the training set differs from that in the test set, remains a critical issue in VQA. Over the past years, numerous debiasing methods have been proposed, which can be broadly categorized into five groups. The first category comprises methods that employ novel model architectures~\cite{11,13,45,54}. The second includes approaches that enhance the language model (LM)~\cite{53,55}. The third focuses on methods that improve visual attention mechanisms~\cite{14,15,90}. The fourth consists of ensemble methods~\cite{56,57,58,59,60}, including recent works that use loss estimation scheme~\cite{89}, adversarial networks~\cite{64}, margin losses~\cite{66}, and causal inference~\cite{69} to diminish bias. The fifth category involves data-driven strategies~\cite{16,17,18,19,20}. 

These contributions have significantly advanced debiasing research. Notably, Zhu \textit{et al.}~\cite{7} introduced a self-supervised debiasing method that pioneered the use of self-supervised strategies in VQA debiasing, and Cao \textit{et al.}~\cite{67} later proposed a robust VQA method based on this approach by integrating contrastive learning and designing a novel loss function. However, none of these methods generate highly irrelevant samples, and as a result, the QI correlation learning mechanisms remain insufficiently effective. In addition, recent work~\cite{68,70,91} on debiasing and robustness for VQA methods using pre-trained large models has pointed out that LLMs, PLMs, and VLPs also endure language bias and lack robustness when handling out-of-distribution data. Given these challenges and the deployment difficulties of large models, we focus on a lightweight yet efficient data strategy for boosting VQA performance. 
In this work, we enhance QI relation learning with a novel generation-based strategy that employs a diffusion model and a sentence revision tool to automatically produce highly contrasting samples. Compared with the methods employing counterfactual learning and causal learning, our method concentrates on the deeper contrastive learning of the visual-text relation, and this feature endows the proposed method with higher compatibility and generalization.

\subsection{Self-supervised Learning}

Self-supervised learning generates supervisory signals directly from the input data, enabling the extraction of high-level representations for various downstream tasks, such as image augmentation~\cite{29}. This paradigm has demonstrated success in representation and visual feature learning within computer vision~\cite{23}. With recent progress in self-supervised and deep learning, these techniques have also been effectively applied to multimodal domains. 
In our work, we propose an innovative self-supervised scheme to further advance debiasing and robustness in VQA.

\section{Proposed Method}
\label{sec:metho}
\subsection{Optimized QI Learning}

QI correlation learning is critical for VQA tasks, as the input QI pairs significantly influence model predictions. Since models tend to produce incorrect answers when provided with non-matching (irrelevant) QI inputs, thereby, we propose an optimized QI correlation learning scheme. To encourage the VQA model to learn authentic QI relations from relevant pairs and prevent it from acquiring superficial correlations from irrelevant ones, we devise a highly negative sample generation strategy that utilizes a sentence revision tool and a diffusion model. As training progresses, the loss functions guide the model to rely on relevant QI pairs, and due to the high irrelevance between the generated questions and images, the effectiveness of QI relation learning is further enhanced. 

\begin{figure*}[!t]
	\centering
	\includegraphics[width = \textwidth]{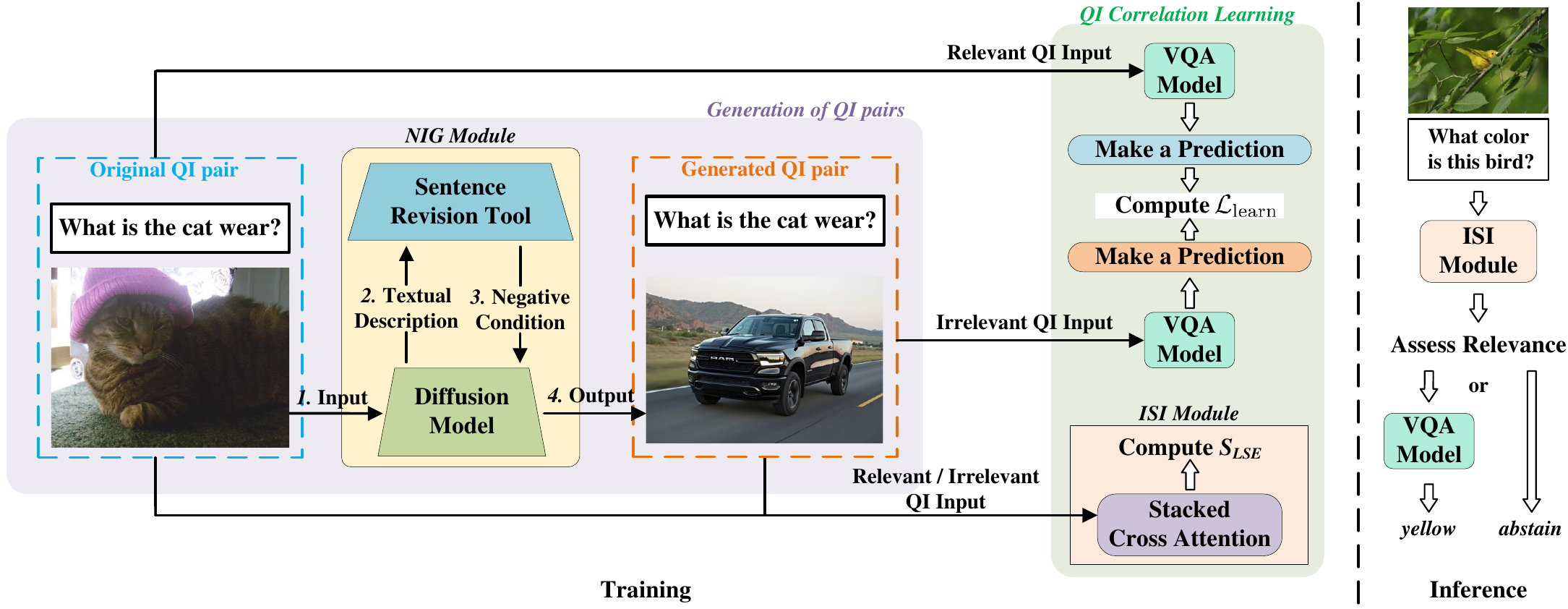}
	
	\caption{\textbf{Proposed QIRL Architecture Comprises Three Components:} a VQA model, a Generation of QI Pairs module, and a QI Correlation Learning module. The VQA model produces predictions based on the inputs; the Generation of QI Pairs process (purple zone) enhances the quality of generated QI pairs by ensuring they are highly irrelevant; and the QI Correlation Learning process (aqua zone) improves VQA performance by alleviating language bias and increasing model capacity.}
	\label{fig4}
\end{figure*}

\subsection{Architecture}
\label{sec: archi}

The architecture of the proposed method is illustrated in \cref{fig4} and comprises three main components: the VQA model, the generation of QI pairs, and QI correlation learning. The VQA model is responsible for making predictions based on the inputs. The second component focuses on improving the quality of generated QI pairs to ensure high irrelevance, while the third component promotes VQA performance by alleviating language bias and enhancing model capacity. These components jointly achieve debiasing and robustness promotion through the integration of the Negative Image Generation (NIG) module and the Irrelevant Sample Identification (ISI) module.

\subsubsection{VQA Model} 

A VQA system, denoted by \textit{F}, takes an image and a free-form, open-ended natural language question as input and produces a natural language answer~\cite{1}. Typically, the VQA problem is formulated as a multi-class classification task. In this formulation, an image \textit{I} is processed by a convolutional neural network (CNN), and a question \textit{Q} is processed by a language module.
The image and question are then jointly mapped to candidate answers \textit{A}~\cite{32}, and the objective is to predict an answer $\hat{a}$ from all candidate answers $a$ based on the image content and question context. Without loss of generality, a VQA model can be formulated as a function transformation:
\begin{equation}\label{eq1}
	\hat{a} = \underset{a\in A}{\operatorname*{\arg \max}} p \left(a \mid Q,I;\theta \right),
\end{equation}
where $\theta$ denotes the model parameters. A common approach to predict the answer is via the cross-entropy loss, as shown in \cref{eq2}, or using a multi-label soft loss, as shown in \cref{eq3}:
\begin{equation}\label{eq2}
	\mathcal{L}_\mathrm{ce} = -\frac{1}{N} \sum_{i = 1}^{N} a_i \log p_i, \quad
	p_i = \mathrm{softmax} \left(W h_i + b \right),
\end{equation}
where $N$ is the number of samples, and $W$ and $b$ are learnable parameters, with $h_i$ representing the fused multi-modal feature.
\begin{equation}\label{eq3}
\begin{aligned}
	\mathcal{L}_\mathrm{ml} = & -\frac{1}{N} \sum_{i = 1}^{N} \bigg[t_i \log \delta \left(F \left(A \mid I_i,Q_i \right) \right) \\
	& + \left(1 - t_i \right) \log \left(1 - \delta \left(F \left(A \mid I_i,Q_i \right) \right) \right) \bigg],
\end{aligned}
\end{equation}
where $\delta(\cdot)$ is the sigmoid function and $t_i$ is the soft target score for candidate answer $a$.

\subsubsection{Generation of QI Pairs}

As illustrated in \cref{fig4}, this component consists of a diffusion model and a sentence simplification tool. Considering the strong cross-modal conversion capabilities of Versatile Diffusion (VD)~\cite{34}, we employ it for both image-to-text and text-to-image generation. To restrict the content of generated images and avoid similarities to the original image, we utilize an unsupervised revision manner~\cite{33} to produce condition sentences. Initially, the original image of the input QI pair is sent to VD to generate a caption. This caption is then converted into a negative caption through sentence editing, which serves as the condition for new image generation. By introducing a negative condition, the newly generated image becomes highly irrelevant to the original question. Compared with existing methods that randomly substitute the original image from the dataset, our process is more accurate and effective. 

\subsubsection{QI Correlation Learning} 

Following the approach in~\cite{7}, this component trains the model to estimate the relevance of QI pairs and encourages it to become independent of superficial correlations, thereby alleviating language bias. Specifically, the original QI pair is labeled as relevant ($c = 1$), while the generated QI pair is labeled as irrelevant ($c = 0$). The loss for QI correlation learning, $\mathcal{L}_\mathrm{learn}$, is defined as:
\begin{equation}\label{eq4}
\begin{aligned}
	\mathcal{L}_\mathrm{learn} = & -\frac{1}{2N} \sum_{i = 1}^{2N} \bigg[c_i \log P (A_i \mid I_i,Q_i) \\
	& + \phi (1 - c_i) \log \big(1 - P (A_i \mid I_i,Q_i) \big) \bigg],
\end{aligned}
\end{equation}
where $\phi$ is a hyper-parameter.

Despite this training scheme reducing reliance on superficial QI correlations, the model may still produce incorrect predictions when faced with irrelevant QI inputs during inference. To address this, 
we introduce the ISI module to enhance the robustness of the VQA model. Using stacked cross attention~\cite{46}, the ISI module estimates the similarity $S_\mathrm{LSE}$ of QI pairs during training and inference, it determines whether the VQA model should provide a prediction or abstain. 

\subsection{Negative Image Generation}
\label{sec: nig}

The NIG module comprises VD~\cite{34} and a sentence revision tool~\cite{33}, where VD is responsible for cross-modal conversion and the sentence revision tool handles sentence paraphrasing.

The basic processes of VD are similar to conventional diffusion models. Specifically, the forward diffusion process $q(x_{T} \mid x_{0})$ degrades the data from $x_0$ to $x_T$ by adding random Gaussian noise. Diffusion models are trained to reverse this process and reconstruct $x_0$ from $x_T$ by removing the noise; this reverse process $p(x_{0} \mid x_{T})$ is defined as a Markov chain~\cite{38} with learned Gaussian transitions starting from $p(x_{T}) = N(x_{T}; 0, I)$, as shown:
\begin{equation}\label{eq5}
\begin{array}{c}
	p_{\theta} \left(x_{0} \mid x_{T} \right) = p \left(x_{T} \right) \prod_{t = 1}^{T} p_{\theta} \left(x_{t-1} \mid x_{t} \right), \\ [5pt]
	p_{\theta} \left(x_{t-1} \mid x_{t} \right) = \mathcal{N} \left(x_{t-1}; \mu \left(x_{t}, t \right), \Sigma \left(x_{t}, t \right) \right).
\end{array}
\end{equation}
VD’s forward diffusion process is similarly defined. The forward process from $x_{0}$ to $x_{T}$ over $T$ steps, with random Gaussian noise added according to a variance schedule $\beta_{1}, \ldots, \beta_{T}$, is given by:
\begin{equation}\label{eq6}
\begin{array}{c}
	q \left(x_{T} \mid x_{0} \right) = \prod_{t = 1}^{T} q \left(x_{t} \mid x_{t-1} \right), \\ [5pt]
	q \left(x_{t} \mid x_{t-1} \right) = \mathcal{N} \left(x_{t}; \sqrt{1-\beta_{t}} x_{t-1}, \beta_{t} I \right).
\end{array}
\end{equation} 
Note that the VD's architecture is multiple flows. Benefiting from its peculiar multi-flow framework, VD can handle various unimodal and multimodal tasks such as image variation, text variation, text-to-image, and image-to-text conversion. To achieve efficient cross-modal conversion and restrict the growth of computational cost, we integrate pre-trained four-flow weights in our framework for image-to-text and text-to-image conversion in NIG module. 

The unsupervised sentence simplification approach~\cite{33} is guided by a scoring function that evaluates fluency, simplicity, and meaning preservation. Initially, candidate sentences are generated through a series of lexical and syntactic operations, and an appropriate scoring function is defined to evaluate them. To produce the contrary sentences, we compute a score for each candidate sentence $\tilde{s}$ as follows:
\begin{equation}\label{eq7}
	f \left(\tilde{s} \right) = \frac{f_\mathrm{LM} \left(\tilde{s} \right)^\alpha}{f_\mathrm{SeI} \left(\tilde{s} \right)^\beta f_\mathrm{SyI} \left(\tilde{s} \right)},
\end{equation} 
where $f_\mathrm{LM}(\tilde{s})^\alpha$ measures language fluency using a probabilistic language model~\cite{44}; $f_\mathrm{SeI}(\tilde{s})^\beta$ measures semantic integrity via cosine similarity; and $f_\mathrm{SyI}(\tilde{s})$ assesses syntactic integrity using root tags. The hyper-parameters $\alpha$ and $\beta$ determine the relative importance of the LM score and semantic integrity, respectively.

\begin{figure}[!t]
	\centering
	\includegraphics[width = \linewidth]{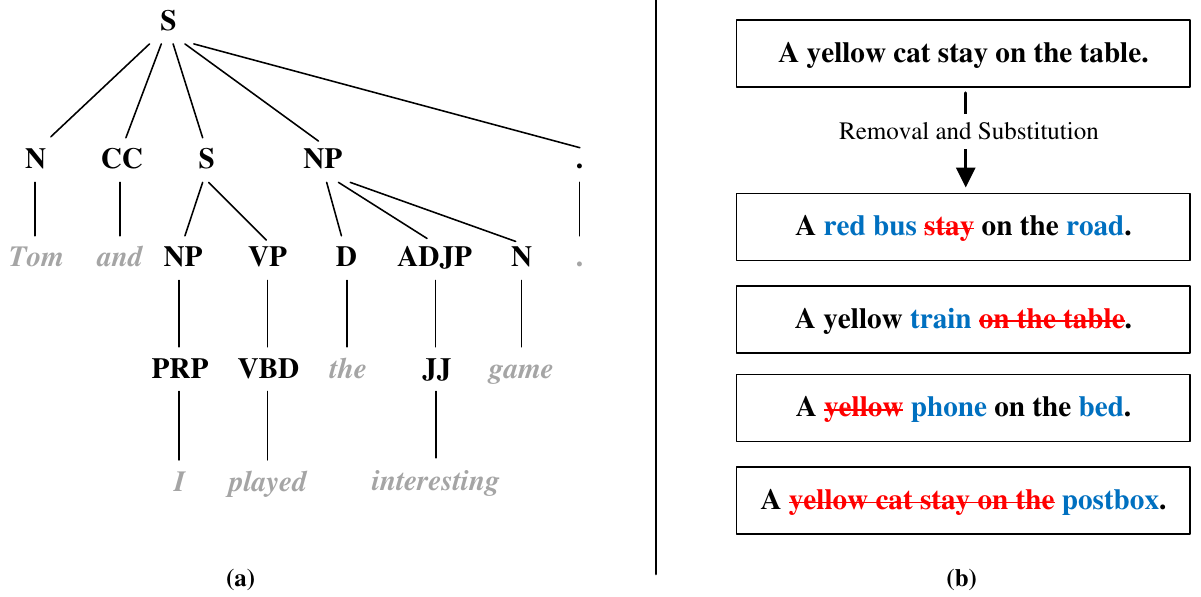}
	
	\caption{\textbf{Sentence Revision Process.} (a) depicts a parse tree for phrase detection; (b) illustrates two operations used to generate candidate sentences, where the blue words and the red words represent the Substitution operation and the Removal operation, respectively.}
	\label{fig5}
\end{figure}

Semantic and syntactic integrity indicate the similarity and preservation of content relative to the original sentence; thus, candidate sentences that are more similar to the original receive lower scores. As illustrated in \cref{fig5}, a constituency parse tree is used to detect phrases (with tags such as $S$, $N$, $CC$, $NP$, $VP$, $PRP$, $VBD$, $D$, $ADJP$, and $JJ$). Unlike the original approach that employs multiple operations (\textit{i.e.}, Removal, Extraction, Reordering, and Substitution), we only perform Removal and Substitution to generate candidate sentences. After sentence editing and score computation, the candidate sentence with the highest score is next selected as the negative condition for VD's image generation.

\begin{figure}[!t]
	\centering
	\includegraphics[width = \linewidth]{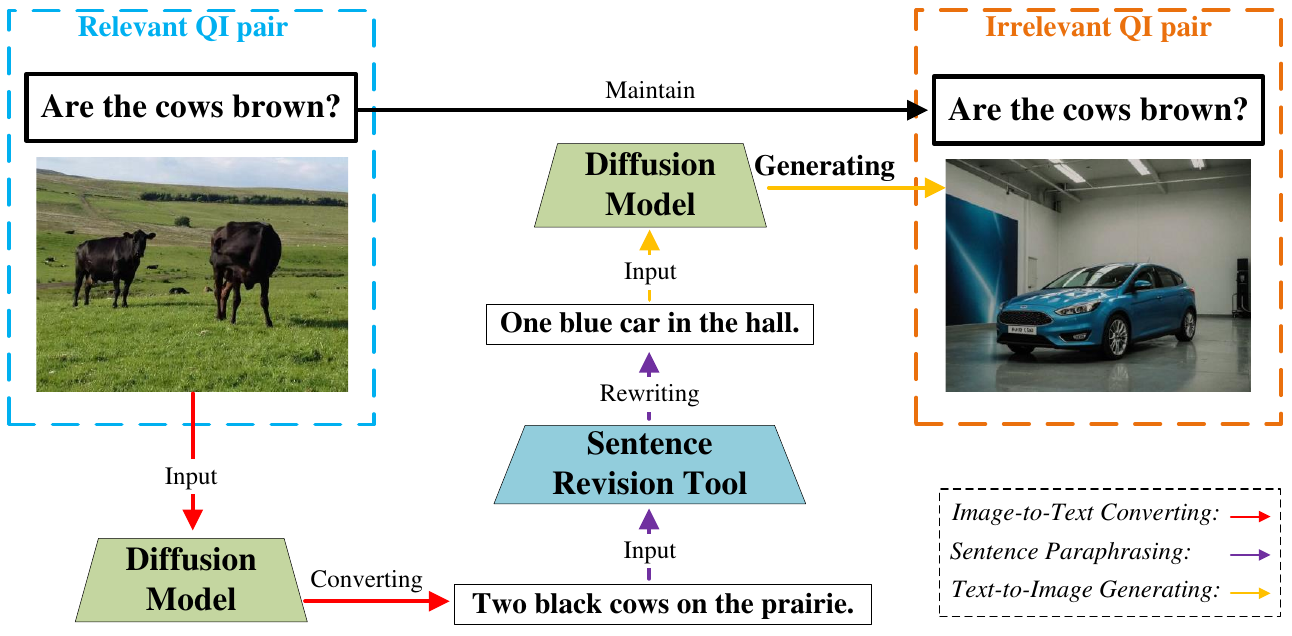}
	
	\caption{\textbf{NIG Procedure.} First, the original image is converted into a caption by VD, which serves as input to the sentence revision module; next, the caption is negatively paraphrased to produce a revised sentence; finally, VD generates a new image based on this revised sentence. This process yields a negative sample comprising the new image paired with the original question.}
	\label{fig6}
\end{figure}

As shown in \cref{fig6}, the NIG module involves three steps:
\textbf{Image-to-Text Conversion:} The original image is converted to a caption by VD, which serves as input to the sentence revision module.
\textbf{Sentence Paraphrasing:} The caption is edited through negative paraphrasing, and the resulting sentence is used for image generation.
\textbf{Text-to-Image Generation:} VD generates a new image based on the edited sentence. After this process, a negative sample is produced, comprising the new image paired with the original question.

\subsection{Irrelevant Sample Identification}
\label{sec: isi}

\begin{figure}[!t]
	\centering
	\includegraphics[width = \linewidth]{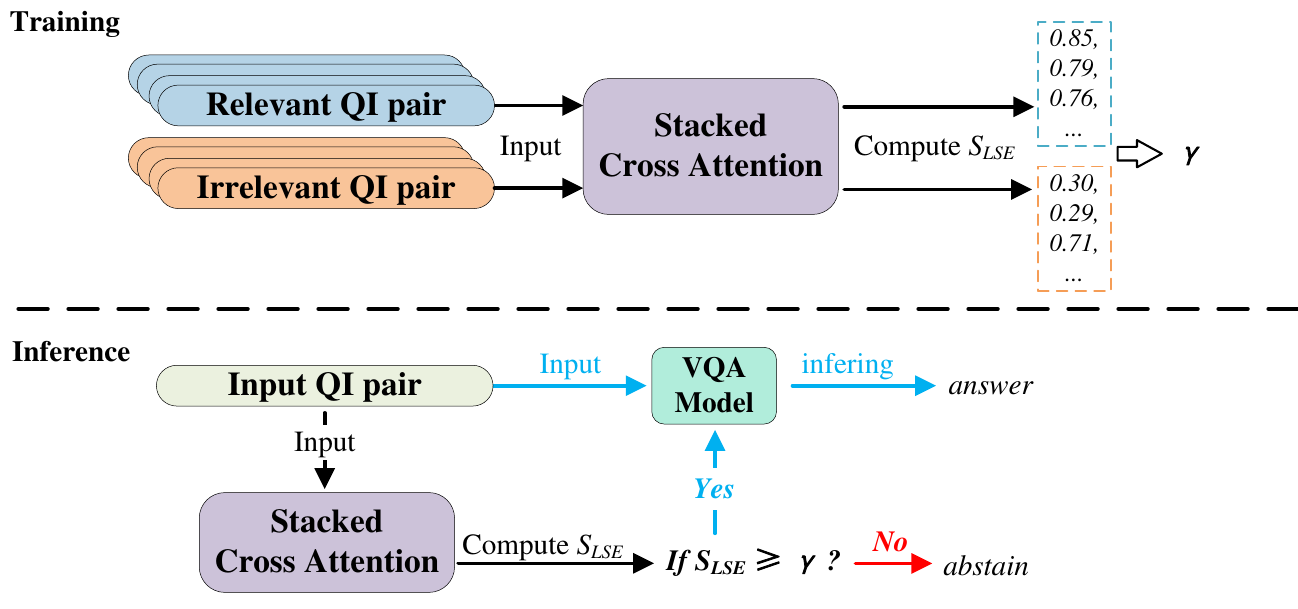}
	
	\caption{\textbf{Illustration of the ISI module.} During training, the module computes the QI similarity, $S_\mathrm{LSE}$, for both relevant and irrelevant QI pairs and determines a threshold $\gamma$. During inference, QI pairs are first evaluated based on their similarity scores; pairs scoring below $\gamma$ are not passed to the VQA model and yield an output of ``abstain'', whereas pairs scoring above $\gamma$ are forwarded to the VQA model for answer prediction.}
	\label{fig7}
\end{figure}

The ISI module functions as a discriminator that determines whether QI inputs are common (relevant) or not. As illustrated in \cref{fig7}, the ISI module is trained jointly with the VQA model during the QI correlation learning phase. During training, it learns to compute a matching degree for the input QI pair and establishes a relation between the relevance label and the matching degree. During inference, the ISI module discriminates whether the input is common or not. As shown in \cref{eq8}, during inference, the ISI module $j$ decides whether the VQA model $f$ should provide an answer or abstain, and the complete decision-making mechanism $o$ is composed of $j$ and $f$:
\begin{equation}\label{eq8}
	o(x) = \left(f,j \right)(x) = \begin{cases} 
	f(x) & \text{if } j(x) = 1, \\
	\emptyset & \text{if } j(x) = 0,
	\end{cases}
\end{equation}
where the discriminator $j(x)$ is defined based on a threshold $\gamma$:
\begin{equation}\label{eq9}
	j(x) = \begin{cases} 
	1 & \text{if } j^{\prime}(x) \geq \gamma; \\
	0 & \text{if } j^{\prime}(x) < \gamma,
	\end{cases}
\end{equation}
and $j^{\prime}(x): x\to[0,1]$ predicts a similarity score that determines whether $f$ should yield an answer.

We use stacked cross attention~\cite{46} to compute the matching degree (relevance level) of QI pairs. The training process consists of three steps: Feature encoding, attention to words, and attention to image regions.

\subsubsection{Feature Encoding} 

The input image $I$ and question $Q$ are encoded into a set of image features $V = (v_1, v_2, \ldots, v_n)$, with each $v_i$ representing an image region, and a set of word features $E = (e_1, e_2, \ldots, e_m)$, where each $e_j$ encodes a word from the question. These features are obtained using UpDn~\cite{45} for the image and an LSTM~\cite{30} for the question. The cosine similarity $s_{ij}$ between each image feature $v_i$ and word feature $e_j$ is computed as:
\begin{equation}\label{eq10}
	s_{ij} = \frac{v_i^T e_j}{\|v_i\| \|e_j\|}, \quad i\in[1,n], j\in[1,m].
\end{equation}

\subsubsection{Attention to Words} 

Attention is imposed on the words for each image region. The attended sentence vector $a^t_i$ for the $i^\text{th}$ image region is defined as:
\begin{equation}\label{eq11}
	\begin{array}{c}
	a^t_i = \sum_{j = 1}^{m} \xi_{ij} e_j, \\ [5pt]
	\xi_{ij} = \frac{\exp \lambda_1 \bar{s}_{ij}}{\sum_{j = 1}^{m} \exp \lambda_1 \bar{s}_{ij}},
	\end{array}
\end{equation} 
where $\lambda_1$ is the inverse temperature parameter of the softmax function, and $\xi_{ij}$ represents a variant of dot product attention.

\subsubsection{Attention to Image Regions} 

The relevance between the $i^\text{th}$ image region and the sentence is measured by the cosine similarity between the image feature $v_i$ and its corresponding attended sentence vector $a^t_i$:
\begin{equation}\label{eq12}
	R \left(v_i,a^t_i \right) = \frac{v_i^T a^t_i}{\|v_i\| \|a^t_i\|}, \quad i\in[1,n].
\end{equation}
The overall similarity between image $I$ and question $Q$ is then calculated using LogSumExp pooling:
\begin{equation}\label{eq13}
	S_\mathrm{LSE} \left(I,Q \right) = \log \left(\sum_{i = 1}^{n} \exp \lambda_2 R \left(v_i,a^t_i \right) \right)^{\frac{1}{\lambda_2}},
\end{equation}
where $\lambda_2$ is a parameter that controls the amplification of the most relevant feature pairs.

The hinge-based triplet ranking loss is used as the objective for image-text matching with margin $\xi$. As shown in \cref{eq14}, the first term sums over all negative questions $\hat{Q}$ for a given image $I$, while the second term sums over all negative images $\hat{I}$ for a given question $Q$. If $I$ and $Q$ are closer in the joint embedding space than any negative pair by a margin $\xi$, the hinge loss becomes zero.
\begin{equation}\label{eq14}
\begin{aligned}
	L \left(I,Q \right) & = \sum_{\hat{Q}} \left[\xi - S_\mathrm{LSE} \left(I,Q \right) + S_\mathrm{LSE} \left(I,\hat{Q} \right) \right]_{+} \\
 & + \sum_{\hat{I}} \left[\xi - S_\mathrm{LSE} \left(I,Q \right) + S_\mathrm{LSE} \left(\hat{I},Q \right) \right]_{+},
\end{aligned}
\end{equation}
where $[x]_+ \equiv \max(x,0)$.

After training, the threshold $\gamma$ is chosen based on the overlapping region of similarity scores for relevant ($c = 1$) and irrelevant ($c = 0$) QI pairs.

\section{Experimental Results}
\label{sec:experi}

\subsection{Environment Setup}

\subsubsection{Datasets and Metrics}

We evaluate the proposed method on two widely used benchmarks: \textsc{VQAv2}~\cite{2} and \textsc{VQA-CPv2}~\cite{11}. \textsc{VQAv2} is extensively used for training and evaluating VQA models and comprises 265,016 images, including both COCO images and abstract scenes. The COCO images provide rich visual information from real-world environments, featuring diverse objects, scenes, and activities that underpin the associated questions and answers. In contrast, \textsc{VQA-CPv2} dataset is derived by rearranging the training and validation splits of \textsc{VQAv2} so that the relation between question types and correct answers differs across the splits. This variation creates a more challenging and realistic evaluation setting, offering insights into the limitations and potential improvements of VQA models.

In our experiments, we employ two metrics. The first is the standard VQA evaluation metric~\cite{1,2}, where each predicted answer $a$ is computed as in \cref{eq15}. In this metric, each question is answered by ten annotators, and the score accounts for discrepancies in human responses.
\begin{equation}\label{eq15}
	\mathrm{Acc} = \min \left(1, \frac{\#\ \text{humans that provide answer } a}{3} \right).
\end{equation}
The second is a specialized metric designed to assess the performance of the ISI module on irrelevant QI pairs. Under this metric, the predicted answer $a$ is computed as in \cref{eq16}:
\begin{equation}\label{eq16}
	\mathrm{Acc}_\mathrm{spe} = \begin{cases} 
	\mathrm{Acc}, & \text{if } a \neq \text{abstain}; \\
	1, & \text{if } a = \text{abstain}.
	\end{cases}
\end{equation}
In other words, while the standard metric treats an ``abstain'' response as incorrect (due to the lack of annotation), the specialized metric regards it as correct, thereby evaluating the selective mechanism's effectiveness.

\subsubsection{Implementation Details}

We follow the implementation and training schedule of previous work~\cite{7}. In our main experiments, we adopt UpDn~\cite{45} and LMH~\cite{56} for VQA inference, and employ the multi-label soft loss as the VQA loss. For visual data processing, we utilize pre-trained four-flow weights. Regarding hyper-parameters, we set $\phi$ in $L_\mathrm{learn}$ to 3; $\alpha$ and $\beta$ in $f(\tilde{s})$ to 0.3 and 1, respectively; and $\lambda_1$ in $\xi_{ij}$ and $\lambda_2$ in $S_\mathrm{LSE}(I,Q)$ to 4 and 5, respectively. During training, the threshold $\gamma$ in $j(x)$ is fixed at 0.71.

\begin{table*}[!t]
	\centering
	\setlength{\tabcolsep}{4pt}
	\caption{\textbf{Performance Comparison on \textsc{VQA-CPv2 test set} and \textsc{VQAv2 validation set}.} Y/N, Num, and Others denote three different question types. The best results are highlighted in bold.}
	\begin{tabular}{ll|cccc|cccc}
	\toprule[1.1pt]
	\multirow{2}[2]{*}{Method} & \multirow{2}[2]{*}{Venue} & \multicolumn{4}{c|}{\textsc{VQA-CPv2}} & \multicolumn{4}{c}{\textsc{VQAv2}} \\
	\cmidrule(lr){3-6} \cmidrule(lr){7-10}
	& & Y/N-CP & Num-CP & Others-CP & Overall-CP & Y/N & Num & Others & Overall \\
	\midrule
	\multicolumn{10}{c}{\textit{Basic Methods}} \\
	\midrule
	SAN~\cite{54} & CVPR'16 & 38.35 & 11.14 & 21.74 & 24.96 & 70.06 & 39.28 & 47.84 & 52.41 \\
	GVQA~\cite{11} & CVPR'18 & 57.99 & 13.68 & 22.14 & 31.30 & 72.03 & 31.17 & 34.65 & 48.24 \\
	UpDn~\cite{45} & CVPR'18 & 42.27 & 11.93 & 46.05 & 39.74 & 81.18 & 42.14 & 55.66 & 63.48 \\
	S-MRL~\cite{13} & NeurIPS'19 & 42.85 & 12.81 & 43.20 & 38.46 & 41.96 & 12.54 & 41.35 & 37.13\\
	\midrule
	\multicolumn{10}{c}{\textit{Methods with Improved LM}} \\
	\midrule
	DLR~\cite{53} & AAAI'20 & 70.99 & 18.72 & 45.57 & 48.87 & 76.82 & 39.33 & 48.54 & 57.96 \\
	VGQE~\cite{55} & ECCV'20 & 66.35 & 27.08 & 46.77 & 50.11 & - & - & - & 64.04 \\
	\midrule
	\multicolumn{10}{c}{\textit{Methods with Enhanced Visual Attention}} \\
	\midrule
	HINT~\cite{14} & ICCV'19 & 67.27 & 10.61 & 45.88 & 46.73 & 81.18 & 42.99 & 55.56 & 63.38 \\
	SCR~\cite{15} & NeurIPS'19 & 70.41 & 10.42 & 47.29 & 48.47 & 78.80 & 41.60 & 54.50 & 62.20 \\
	AttReg~\cite{90} & TOMM'22 & 86.80 & 51.37 & 48.25 & 60.00 & 79.66 & 41.68 & 55.42 & 62.63 \\
	\midrule
	\multicolumn{10}{c}{\textit{Ensemble Methods}} \\
	\midrule
	AdvReg~\cite{12} & NeurIPS'18 & 65.49 & 15.48 & 35.48 & 41.17 & 79.84 & 42.35 & 55.16 & 62.75 \\
	RUBi~\cite{13} & NeurIPS'19 & 68.65 & 20.28 & 43.18 & 47.11 & - & - & - & - \\
	LMH~\cite{56} & EMNLP'19 & 70.29 & 44.10 & 44.86 & 52.15 & 65.06 & 37.63 & 54.69 & 56.35 \\
	AdaVQA~\cite{60} & IJCAI'21 & 70.83 & 49.00 & 46.29 & 54.02 & 47.78 & 34.13 & 51.14 & 46.98 \\
	CF-VQA~\cite{57} & CVPR'21 & 90.61 & 21.50 & 45.61 & 55.05 & 81.13 & 43.86 & 50.11 & 60.94 \\
	IntroD~\cite{58} & NeurIPS'21 & \textbf{90.79} & 17.92 & 46.73 & 55.17 & 82.48 & \textbf{46.60} & 54.05 & 63.40 \\
	GGE~\cite{59} & ICCV'21 & 87.04 & 27.75 & 49.59 & 57.32 & 73.27 & 39.99 & 54.39 & 59.11 \\
	CI-VQA~\cite{89} & TIP'22 & 72.78 & 48.00 & 44.31 & 53.22 & 68.16 & 36.35 & 51.26 & 56.70 \\
	GenB~\cite{64} & CVPR'23 & 88.03 & 40.05 & 49.25 & 59.15 & - & - & - & - \\
	RMLVQA~\cite{66} & CVPR'23 & 89.98 & 45.96 & 48.74 & 60.41 & 76.68 & 37.54 & 53.26 & 59.99 \\
	DCCE~\cite{69} & EMNLP'24 & 90.66 & 45.73 & 46.03 & 58.70 & 77.51 & 40.26 & 53.16 & 60.43 \\
	\midrule
	\multicolumn{10}{c}{\textit{Methods with Data Strategy}} \\
	\midrule
	{LMH~\cite{56} \textit{w/} CSS~\cite{17}} & CVPR'20 & 84.37 & 49.42 & 48.21 & 58.95 & 73.25 & 39.77 & 55.11 & 59.91 \\
	{LMH~\cite{56} \textit{w/} CSSCL~\cite{19}} & EMNLP'20 & 86.99 & 49.89 & 47.16 & 59.18 & 67.27 & 38.40 & 54.71 & 57.29 \\
	\rowcolor{gray!20}
	{LMH~\cite{56} \textit{w/} QIRL} (Ours) & & \begin{tabular}{c} 88.22 \\ ($\uparrow$ 17.93) \end{tabular} & \begin{tabular}{c} 51.07 \\ ($\uparrow$ 6.97) \end{tabular} & \begin{tabular}{c} 50.68 \\ ($\uparrow$ 5.82) \end{tabular} & \begin{tabular}{c} 60.03 \\ ($\uparrow$ 7.88) \end{tabular} & \begin{tabular}{c} 82.52 \\ ($\uparrow$ 17.46) \end{tabular} & \begin{tabular}{c} 45.66 \\ ($\uparrow$ 8.03) \end{tabular} & \begin{tabular}{c} 57.42 \\ ($\uparrow$ 2.73) \end{tabular} & \begin{tabular}{c} 64.16 \\ ($\uparrow$ 7.81) \end{tabular} \\
	\midrule
	{UpDn~\cite{45} \textit{w/} CVL~\cite{16}} & CVPR'20 & 45.72 & 12.45 & 48.34 & 42.12 & - & - & - & - \\
	{UpDn~\cite{45} \textit{w/} RandImg~\cite{65}} & NeurIPS'20 & 83.89 & 41.60 & 44.20 & 55.37 & 76.53 & 33.87 & 48.57 & 57.24 \\
	{UpDn~\cite{45} \textit{w/} SSL~\cite{7}} & IJCAI'20 & {86.53} & {29.87} & {50.03} & {57.59} & - & - & - & {63.73} \\
	{UpDn~\cite{45} \textit{w/} MUTANT~\cite{18}} & EMNLP'20 & 88.90 & 49.68 & 50.78 & 61.72 & 82.07 & 42.52 & 53.28 & 62.56 \\
	{UpDn~\cite{45} \textit{w/} Unshuffling~\cite{61}} & ICCV'21 & 47.72 & 14.43 & 47.24 & 42.39 & 78.32 & 42.16 & 52.81 & 61.08 \\
	{UpDn~\cite{45} \textit{w/} D-VQA~\cite{20}} & NeurIPS'21 & 88.93 & 52.12 & 50.30 & \textbf{61.82} & 82.11 & 44.01 & 57.40 & 63.69 \\
	{UpDn~\cite{45} \textit{w/} KDDAug~\cite{63}} & ECCV'22 & 86.13 & \textbf{55.08} & 48.08 & 60.24 & 80.55 & 41.05 & 55.18 & 62.86 \\
	\rowcolor{gray!20}
	{UpDn~\cite{45} \textit{w/} QIRL} (Ours) & & \begin{tabular}{c} 88.91 \\ ($\uparrow$ 46.64) \end{tabular} & \begin{tabular}{c} 51.60 \\ ($\uparrow$ 39.67) \end{tabular} & \begin{tabular}{c} \textbf{51.44} \\ ($\uparrow$ 5.39) \end{tabular} & \begin{tabular}{c} 61.73 \\ ($\uparrow$ 21.99) \end{tabular} & \begin{tabular}{c} \textbf{82.75} \\ ($\uparrow$ 1.57) \end{tabular} & \begin{tabular}{c} 46.40 \\ ($\uparrow$ 4.26) \end{tabular} & \begin{tabular}{c} \textbf{57.69} \\ ($\uparrow$ 2.03) \end{tabular} & \begin{tabular}{c} \textbf{64.92} \\ ($\uparrow$ 1.44) \end{tabular} \\
	\bottomrule[1.1pt]
	\end{tabular}%
	\label{tab1}%
\end{table*}

\begin{figure*}[!t]
	\centering
	\includegraphics[width = \linewidth]{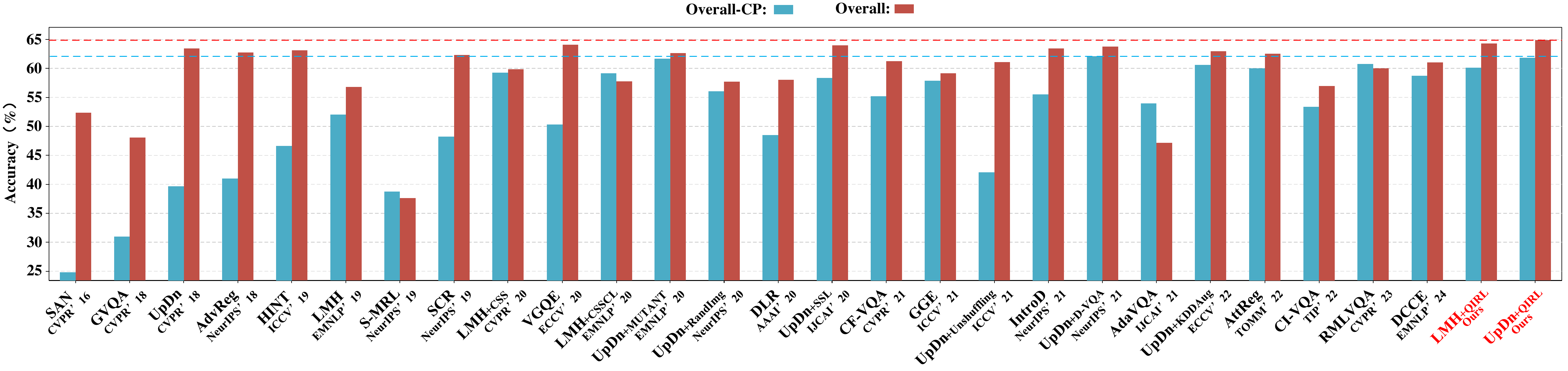}
	
	\caption{\textbf{Quantitative Analysis of Debiasing Studies.} Blue dashed line and red dashed line in the figure indicate the best result in overall accuracy of \textsc{VQA-CPv2 test set} and \textsc{VQAv2 validation set}, respectively.}
	\label{fig8}
\end{figure*}

\subsection{Quantitative Results}

The main quantitative results are summarized in \cref{tab1} and \cref{fig8}. Columns 3 to 6 report results on \textsc{VQA-CPv2} test set, while columns 7 to 10 present results on \textsc{VQAv2} validation set. We report accuracies for various question types, namely, yes/no (Y/N), number (Num), and other (Others), as well as overall accuracy.
 
\textbf{Compared with baseline models,} our method yields remarkable improvements. For the LMH model~\cite{56} on \textsc{VQA-CPv2} test set, our approach improves yes/no accuracy by 17.93\%, number accuracy by 6.97\%, and other accuracy by 5.82\%, resulting in an overall accuracy gain of 7.88\% on \textsc{VQA-CPv2} and 7.81\% on \textsc{VQAv2}. For the UpDn model~\cite{45}, our method enhances yes/no, number, and other accuracies on \textsc{VQA-CPv2} test set by 46.64\%, 39.67\%, and 5.39\%, respectively, leading to overall accuracy gains of 21.99\% on \textsc{VQA-CPv2} and 1.44\% on \textsc{VQAv2}. 

\textbf{Compared with methods based on data strategies,} our approach achieves competitive performance across various question types on \textsc{VQA-CPv2} test set, ranking second in overall accuracy on \textsc{VQA-CPv2} and first on \textsc{VQAv2}. Note that QIRL's results on \textsc{VQA-CPv2} are inferior to those of D-VQA~\cite{20}, which mitigates both language and vision biases. Although D-VQA achieves superior performance, its sophisticated architecture incurs higher training and inference time. Overall, our proposed method outperforms most existing approaches, particularly in overall accuracy, where it attains the best results on both \textsc{VQA-CPv2} test set and \textsc{VQAv2} validation set.

\begin{figure}[!t]
	\centering
	\includegraphics[width = \linewidth]{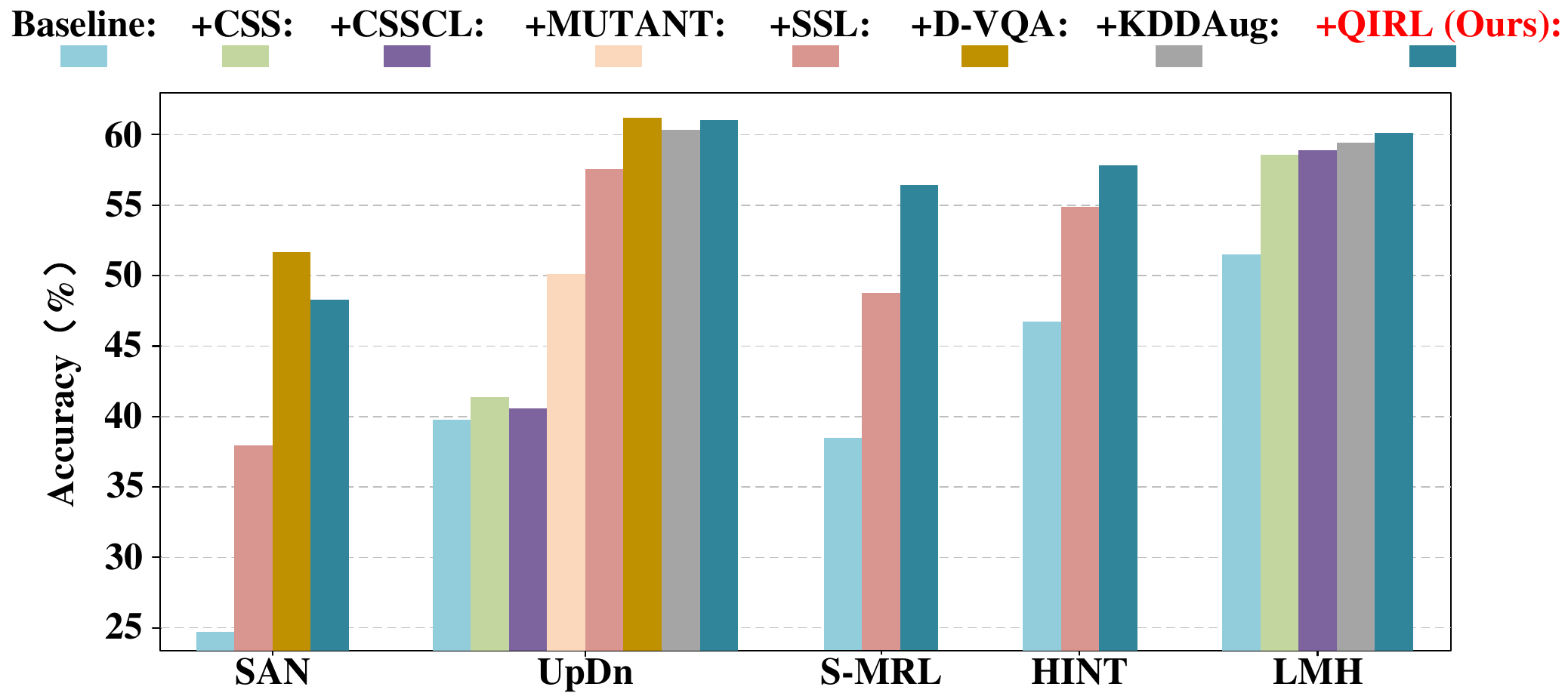}
	
	\caption{\textbf{Performance Comparison of Data Strategy Methods with Different VQA models.} The methods are illustrated with colored bars.}
	\label{fig9}
\end{figure}

\begin{table}[!t]
	\centering
	\setlength{\tabcolsep}{10pt}
	
	\caption{\textbf{Performance Evaluation of Different VQA Baselines on \textsc{VQA-CPv2 test set}.}}
	\begin{tabular}{l|cc}
	\toprule[1.1pt]
	Method & Accuracy & Improvement \\
	\midrule
	SAN~\cite{54} & 24.96 & -- \\
	SAN~\cite{54} \textit{w/} SSL~\cite{7} & 37.64 & $\uparrow$ 12.68 \\
	SAN~\cite{54} \textit{w/} D-VQA~\cite{20} & 52.17 & $\uparrow$ \textbf{27.21} \\
	\rowcolor{gray!20}
	SAN~\cite{54} \textit{w/} QIRL (Ours) & 48.11 & $\uparrow$ 23.15 \\
	\midrule
	UpDn~\cite{45} & 39.74 & -- \\
	UpDn~\cite{45} \textit{w/} CSS~\cite{17} & 41.16 & $\uparrow$ 1.42 \\
	UpDn~\cite{45} \textit{w/} CSSCL~\cite{19} & 40.38 & $\uparrow$ 0.64 \\
	UpDn~\cite{45} \textit{w/} MUTANT~\cite{18} & 50.03 & $\uparrow$ 10.29 \\
	UpDn~\cite{45} \textit{w/} SSL~\cite{7} & 57.59 & $\uparrow$ 17.85 \\
	UpDn~\cite{45} \textit{w/} D-VQA~\cite{20} & 61.82 & $\uparrow$ \textbf{22.08} \\
	UpDn~\cite{45} \textit{w/} KDDAug~\cite{63} & 60.24 & $\uparrow$ 20.50 \\
	\rowcolor{gray!20}
	UpDn~\cite{45} \textit{w/} QIRL (Ours) & 61.73 & $\uparrow$ 21.99 \\
	\midrule
	S-MRL~\cite{13} & 38.46 & -- \\
	S-MRL~\cite{13} \textit{w/} SSL~\cite{7} & 49.09 & $\uparrow$ 10.63 \\
	\rowcolor{gray!20}
	S-MRL~\cite{13} \textit{w/} QIRL (Ours) & 56.14 & $\uparrow$ \textbf{17.68} \\
	\midrule
	HINT~\cite{14} & 46.73 & -- \\
	HINT~\cite{14} \textit{w/} SSL~\cite{7} & 54.96 & $\uparrow$ 8.23 \\
	\rowcolor{gray!20}
	HINT~\cite{14} \textit{w/} QIRL (Ours) & 58.01 & $\uparrow$ \textbf{11.28} \\
	\midrule
	LMH~\cite{56} & 52.15 & -- \\
	LMH~\cite{56} \textit{w/} CSS~\cite{17} & 58.95 & $\uparrow$ 6.80 \\
	LMH~\cite{56} \textit{w/} CSSCL~\cite{19} & 59.18 & $\uparrow$ 7.03 \\
	LMH~\cite{56} \textit{w/} KDDAug~\cite{63} & 59.54 & $\uparrow$ 7.39 \\
	\rowcolor{gray!20}
	LMH~\cite{56} \textit{w/} QIRL (Ours) & 60.03 & $\uparrow$ \textbf{7.88} \\
	\bottomrule[1.1pt]
	\end{tabular}
	\label{tab2}
\end{table}

Further, the results from different VQA baselines on \textsc{VQA-CPv2} test set are presented in \cref{fig9} and \cref{tab2} (with bold fonts indicating the best results). Although D-VQA achieves the best improvements for SAN and UpDn, its approach does not generalize as well to other baselines. In contrast, our method is model-agnostic and can be integrated with various VQA models. Specifically, QIRL improves accuracy by 23.15\% for SAN, 21.99\% for UpDn, 17.68\% for S-MRL, 11.28\% for HINT, and 7.88\% for LMH. In summary, QIRL achieves competitive results compared with other data strategies, realizing the best or second-best improvement for each VQA baseline.

\begin{table*}[!t]
	\centering
	\setlength{\tabcolsep}{9pt}
	
	\caption{\textbf{Performance Comparison of Data Strategy Methods Augmented with the ISI Module on \textsc{VQA-CPv2 test set} and \textsc{VQAv2 validation set}.} Unshaded cells denote results from methods without the ISI module evaluated using the standard metric, while gray cells indicate results from methods with the ISI module evaluated using the specialized metric.}
	\begin{tabular}{lcc|cccc|c}
	\toprule[1.1pt]
	\multirow{2}[2]{*}{Baseline} & \multirow{2}[2]{*}{Venue} & \multirow{2}[2]{*}{Method} & \multicolumn{4}{c|}{\textsc{VQA-CPv2}} & {\textsc{VQAv2}} \\
	\cmidrule(lr){4-7} \cmidrule(lr){8-8}
	& & & Y/N-CP & Num-CP & Others-CP & Overall-CP & Overall \\
	\midrule
	\multirow{2}{*}{RandImg~\cite{65}} & \multirow{2}{*}{NeurIPS'20} & Original & 83.89 & 41.60 & 44.20 & 55.37 & 57.24 \\
	& & \textit{w/} ISI & \cellcolor{gray!20}85.13 ($\uparrow$ 1.24) & \cellcolor{gray!20}44.07 ($\uparrow$ 2.47) & \cellcolor{gray!20}46.77 ($\uparrow$ 2.57) & \cellcolor{gray!20}57.55 ($\uparrow$ 2.18) & \cellcolor{gray!20}60.41 ($\uparrow$ 3.17) \\
	\midrule
	\multirow{2}{*}{CSSCL~\cite{19}} & \multirow{2}{*}{EMNLP'20} & Original & 86.99 & 49.89 & 47.16 & 59.18 & 57.29 \\
	& & \textit{w/} ISI & \cellcolor{gray!20}89.35 ($\uparrow$ 2.36) & \cellcolor{gray!20}51.82 ($\uparrow$ 1.93) & \cellcolor{gray!20}49.30 ($\uparrow$ 2.14) & \cellcolor{gray!20}61.52 ($\uparrow$ 2.34) & \cellcolor{gray!20}59.32 ($\uparrow$ 2.03) \\
	\midrule
	\multirow{2}{*}{MUTANT~\cite{18}} & \multirow{2}{*}{EMNLP'20} & Original & 88.90 & 49.68 & 50.78 & 61.72 & 62.56 \\
	& & \textit{w/} ISI & \cellcolor{gray!20}90.58 ($\uparrow$ 1.68) & \cellcolor{gray!20}51.99 ($\uparrow$ 2.31) & \cellcolor{gray!20}52.77 ($\uparrow$ 1.99) & \cellcolor{gray!20}63.91 ($\uparrow$ 2.19) & \cellcolor{gray!20}65.00 ($\uparrow$ 2.44) \\
	\midrule
	\multirow{2}{*}{CSS~\cite{17}} & \multirow{2}{*}{CVPR'20} & Original & 84.37 & 49.42 & 48.21 & 58.95 & 59.91 \\
	& & \textit{w/} ISI & \cellcolor{gray!20}86.10 ($\uparrow$ 1.73) & \cellcolor{gray!20}52.24 ($\uparrow$ 2.82) & \cellcolor{gray!20}51.07 ($\uparrow$ 2.86) & \cellcolor{gray!20}61.02 ($\uparrow$ 2.07) & \cellcolor{gray!20}61.71 ($\uparrow$ 1.80) \\
	\midrule
	\multirow{2}{*}{SSL~\cite{7}} & \multirow{2}{*}{IJCAI'20} & Original & 86.53 & 29.87 & 50.03 & 57.59 & 63.73 \\
	& & \textit{w/} ISI & \cellcolor{gray!20}88.05 ($\uparrow$ 1.52) & \cellcolor{gray!20}31.66 ($\uparrow$ 1.79) & \cellcolor{gray!20}53.11 ($\uparrow$ 3.08) & \cellcolor{gray!20}59.40 ($\uparrow$ 1.81) & \cellcolor{gray!20}65.51 ($\uparrow$ 1.78) \\
	\midrule
	\multirow{2}{*}{Unshuffling~\cite{61}} & \multirow{2}{*}{ICCV'21} & Original & 47.72 & 14.43 & 47.24 & 42.39 & 61.08 \\
	& & \textit{w/} ISI & \cellcolor{gray!20}49.58 ($\uparrow$ 1.86) & \cellcolor{gray!20}15.97 ($\uparrow$ 1.54) & \cellcolor{gray!20}50.10 ($\uparrow$ 2.86) & \cellcolor{gray!20}45.06 ($\uparrow$ 2.67) & \cellcolor{gray!20}63.35 ($\uparrow$ 2.27) \\
	\midrule
	\multirow{2}{*}{KDDAug~\cite{63}} & \multirow{2}{*}{ECCV'22} & Original & 86.13 & 55.08 & 48.08 & 60.24 & 62.86 \\
	& & \textit{w/} ISI & \cellcolor{gray!20}88.69 ($\uparrow$ 2.56) & \cellcolor{gray!20}58.05 ($\uparrow$ 2.97) & \cellcolor{gray!20}51.12 ($\uparrow$ 3.04) & \cellcolor{gray!20}63.22 ($\uparrow$ 2.98) & \cellcolor{gray!20}64.30 ($\uparrow$ 1.44) \\
	\bottomrule[1.1pt]
	\end{tabular}%
	\label{tab3}%
\end{table*}

\begin{figure}[!t]
	\centering
	\includegraphics[width = \linewidth]{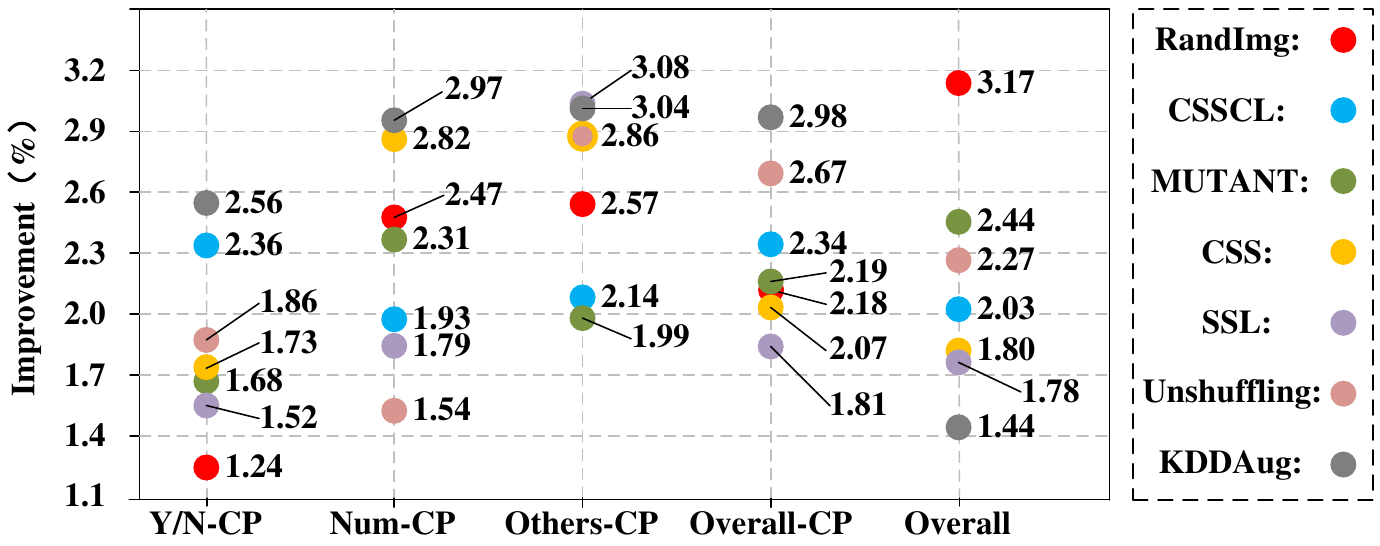}
	
	\caption{\textbf{Evaluation of the Effectiveness of the ISI Module.} Improvements achieved by the ISI module across various methods are illustrated using different colors.}
	\label{fig10}
\end{figure}

Moreover, to further validate the ISI module's effectiveness, we integrate the pre-trained ISI module into other data strategy methods and conduct experiments on \textsc{VQA-CPv2} test set and \textsc{VQAv2} validation set. The results, reported in \cref{tab3} and illustrated in \cref{fig10}, indicate that all methods benefit from the integration of the ISI module, with improvements ranging from at least 1.24\% up to 3.17\% (as observed for RandImg \textit{w/} ISI on \textsc{VQAv2} validation set). These findings clearly demonstrate that the ISI module significantly enhances model robustness for data strategy-based methods.

\begin{figure*}[!h]
	\centering
	\includegraphics[width = 0.8\linewidth]{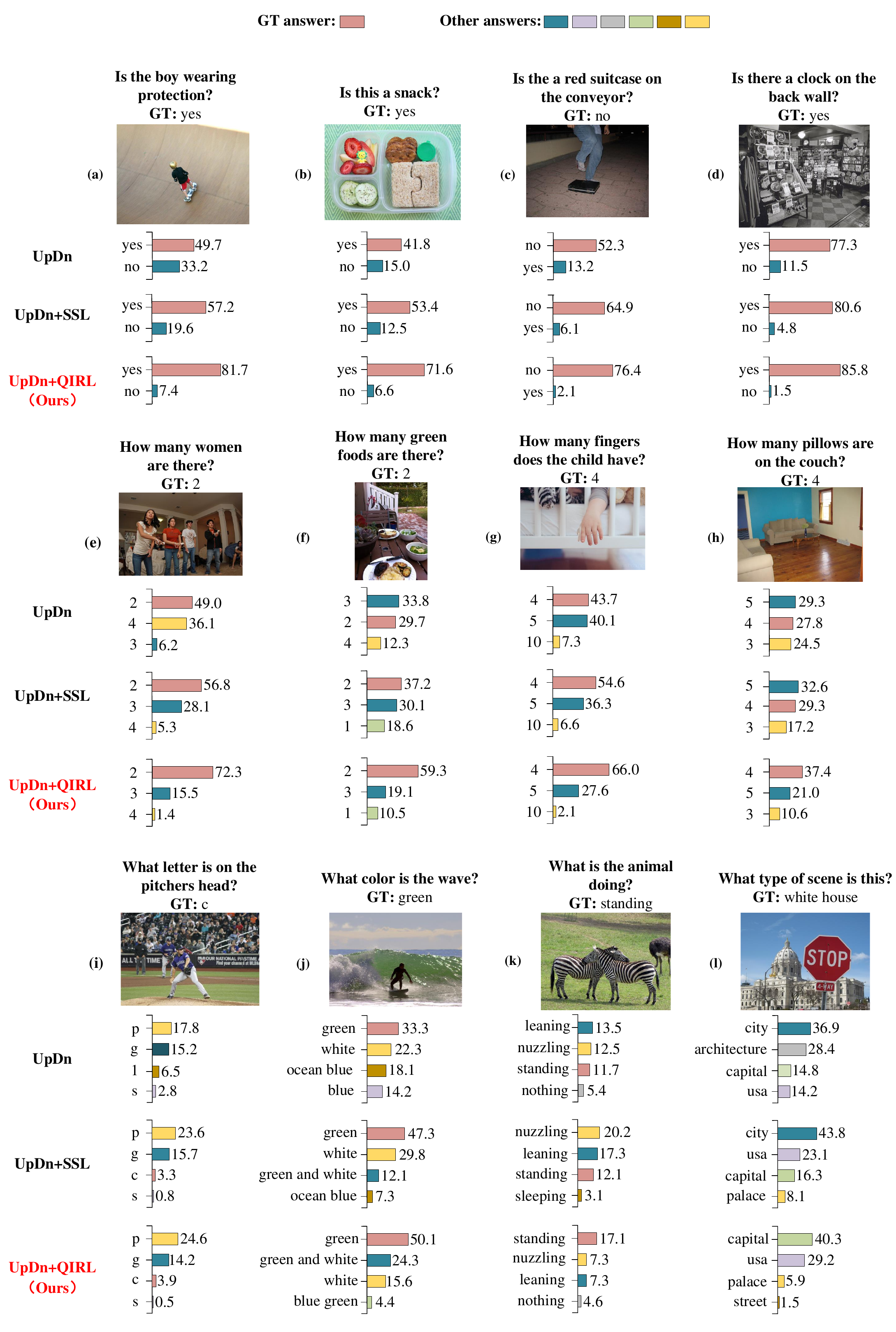}
	\caption{\textbf{Qualitative Analysis.} The red bar means the probability of the GT answer, and other colored bars represent the probability of others.}
	\label{fig11}
\end{figure*}

\subsection{Qualitative Analysis}

\cref{fig11} compares prediction probabilities among the base model, SSL, and our method. 
Cases (a)-(d) correspond to yes/no questions, (e)-(h) to number questions, and (i)-(l) to other questions. 
In each figure, the red bar indicates the probability of the ground-truth answer, while other colors denote competing answers. 
Our method consistently increases the probability of the correct answer, particularly for yes/no questions. 
For number and other question types (\textit{e.g.}, cases (f), (h), (j), and (k)), our model assigns the highest probability to the correct answer while suppressing the influence of distractors. 
Although in cases (i) and (l) none of the models predict the correct answer with the highest probability, due to ambiguous visual cues or fuzzy target regions, our method still demonstrates the strongest relative performance compared to both the base and SSL models.

\subsection{Ablation Studies}

\begin{figure}[!t]
	\centering
	\includegraphics[width = 1.0\linewidth]{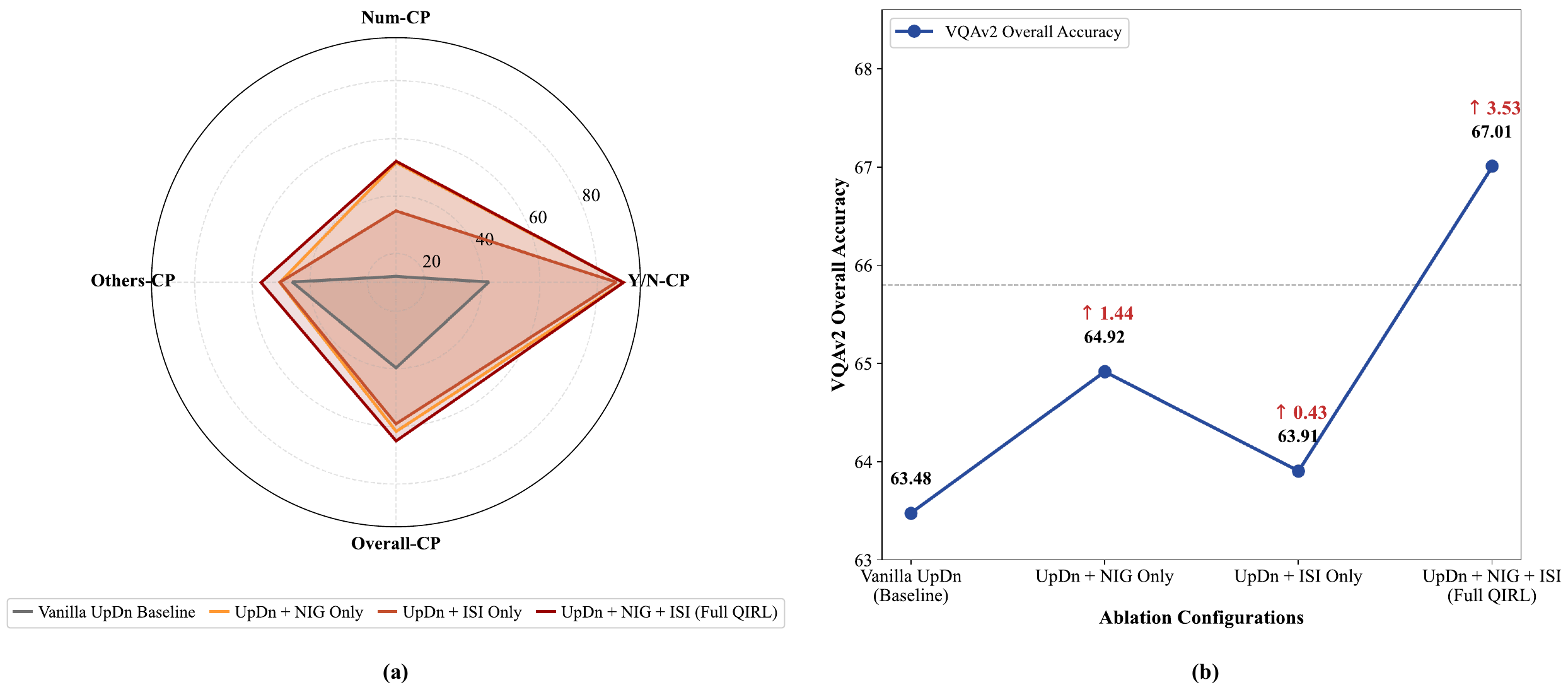}
	\caption{\textbf{Ablation Study on the NIG and ISI modules.} (a) The performance improvement with two modules on \textsc{VQA-CPv2} dataset; (b) The Overall accuracy improvement with two modules on \textsc{VQAv2} dataset.}
	\label{fig12}
\end{figure}

To assess the individual contributions of the NIG and ISI modules, we conduct ablation experiments with and without each component. 
Results based on the specialized metric are shown in \cref{tab4} and \cref{fig12}. 
We consider three variants: 
\textit{1)} with only the NIG module, 
\textit{2)} with only the ISI module (using SSL-based sample generation), and 
\textit{3)} with both modules integrated. 
The variant that uses only NIG produces no ``abstain'' outputs, yielding results identical to those evaluated with the standard metric. 
These experiments demonstrate that both NIG and ISI contribute to performance gains and jointly mitigate language bias in VQA.

\begin{table}[!t]
	\centering
	\setlength{\tabcolsep}{2.5pt} 
	
	\caption{\textbf{Ablation Studies on the NIG and ISI Modules on \textsc{VQA-CPv2 test set} and \textsc{VQAv2 validation set}.}}
	\begin{tabular}{lcc|cccc|c}
	\toprule[1.1pt]
	\multirow{2}[2]{*}{Method} & \multirow{2}[2]{*}{NIG} & \multirow{2}[2]{*}{ISI} & \multicolumn{4}{c|}{\textsc{VQA-CPv2}} & {\textsc{VQAv2}} \\
	\cmidrule(lr){4-7} \cmidrule(lr){8-8}
	& & & Y/N-CP & Num-CP & Others-CP & Overall-CP & Overall \\
	\midrule
	UpDn~\cite{45} & \scalebox{2}{$\circ$} & \scalebox{2}{$\circ$} & 42.27 & 11.93 & 46.05 & 39.74 & 63.48 \\
	{\textit{w/} QIRL} & \scalebox{2}{$\bullet$} & \scalebox{2}{$\circ$} & 88.91 & 51.60 & 50.44 & 61.73 & 64.92 \\
	{\textit{w/} QIRL} & \scalebox{2}{$\circ$} & \scalebox{2}{$\bullet$} & 86.55 & 34.72 & 50.29 & 59.10 & 63.91 \\
	\rowcolor{gray!20}
	{\textit{w/} QIRL} & \scalebox{2}{$\bullet$} & \scalebox{2}{$\bullet$} & 88.93 & 52.14 & 56.93 & 65.03 & 67.01 \\
	\bottomrule[1.1pt]
	\end{tabular}%
	\label{tab4}%
\end{table}

\begin{table*}[!t]
	\centering
	\setlength{\tabcolsep}{10pt}
	\caption{\textbf{Performance Evaluation of Different VQA Losses on \textsc{VQA-CPv2 test set}.}}
	\begin{tabular}{clc|cccc}
	\toprule[1.1pt]
	Loss & Method & Data strategy & Y/N & Num & Others & Overall \\
	\midrule
	\multirow{2}{*}{\begin{tabular}{c} Cross-Entropy \\ Loss \end{tabular}} 
	& UpDn~\cite{45} & No & 47.27 & 13.67 & 40.32 & 38.28 \\
	& {\cellcolor{gray!20}UpDn~\cite{45} \textit{w/} QIRL (Ours)} 
	& {\cellcolor{gray!20}Yes} 
	& {\cellcolor{gray!20}88.92} 
	& {\cellcolor{gray!20}50.28} 
	& {\cellcolor{gray!20}48.70} 
	& {\cellcolor{gray!20}59.20} \\
	\midrule
	\multirow{2}{*}{\begin{tabular}{c} Multi-Label \\ Loss \end{tabular}}
	& UpDn~\cite{45} & No & 43.45 & 13.64 & 48.18 & 41.53 \\
	 & {\cellcolor{gray!20}UpDn~\cite{45} \textit{w/} QIRL (Ours)} 
	& {\cellcolor{gray!20}Yes} 
	& {\cellcolor{gray!20}88.91} 
	& {\cellcolor{gray!20}51.60} 
	& {\cellcolor{gray!20}51.44} 
	& {\cellcolor{gray!20}61.73} \\
	\bottomrule[1.1pt]
	\end{tabular}%
	\label{tab5}%
\end{table*}

To assess the performance under different VQA losses, the performance of our method with different VQA loss functions on \textsc{VQA-CPv2} test set is shown in \cref{tab5}. 
We report results using both cross-entropy and multi-label losses. 
With cross-entropy loss, QIRL achieves gains of 1.17\%, 23.88\%, and 7.28\% for different question types, and an overall accuracy improvement of 6.57\%. 
With multi-label loss, the respective improvements are 2.38\%, 21.73\%, and 0.41\%, yielding a 4.14\% overall gain.

\begin{table}[!t]
	\centering
	\setlength{\tabcolsep}{3.8pt}
	
	\caption{\textbf{Comparison of Standard ($\mathrm{Acc}$) and Specialized ($\mathrm{Acc_{spe}}$) Metrics on \textsc{VQA-CPv2 test set} and \textsc{VQAv2 validation set}.}}
	\begin{tabular}{l|cccc|c}
	\toprule[1.1pt]
	\multirow{2}[2]{*}{Method} & \multicolumn{4}{c|}{\textsc{VQA-CPv2}} & {\textsc{VQAv2}} \\
	\cmidrule(lr){2-5} \cmidrule(lr){6-6}
	& Y/N-CP & Num-CP & Others-CP & Overall-CP & Overall \\
	\midrule
	QIRL -- $\mathrm{Acc}$ & 88.91 & 51.60 & 51.44 & 61.73 & 64.92 \\
	\rowcolor{gray!20}
	QIRL -- $\mathrm{Acc_{spe}}$ & 88.93 & 52.14 & 56.93 & 65.03 & 67.01 \\
	\bottomrule[1.1pt]
	\end{tabular}%
	\label{tab6}%
\end{table}

\begin{figure}[!t]
	\centering
	\includegraphics[width = 0.5\linewidth]{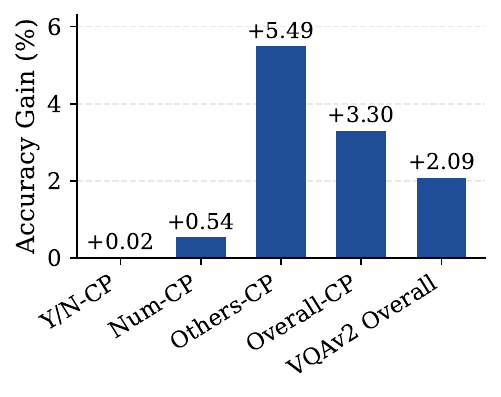}
	\caption{\textbf{Comparison between standard ($\mathrm{Acc}$) and specialized ($\mathrm{Acc_{spe}}$) metrics.} Numbers marked with a plus sign represent the increase in accuracy under the specialized metrics evaluation compared to the standard evaluation metrics.}
	\label{fig13}
\end{figure}

To evaluate robustness against irrelevant QI pairs, we further use the specialized metric that treats ``abstain'' outputs as correct~\citep{1}. 
Evaluation on \textsc{VQA-CPv2} validation and \textsc{VQA-v2} test sets (\cref{tab6,fig13}) shows substantial improvements when ``abstain'' responses are counted as correct, confirming the efficacy of the proposed selective inference mechanism.

\begin{figure}[!t]
	\centering
	\includegraphics[width = 0.8\linewidth]{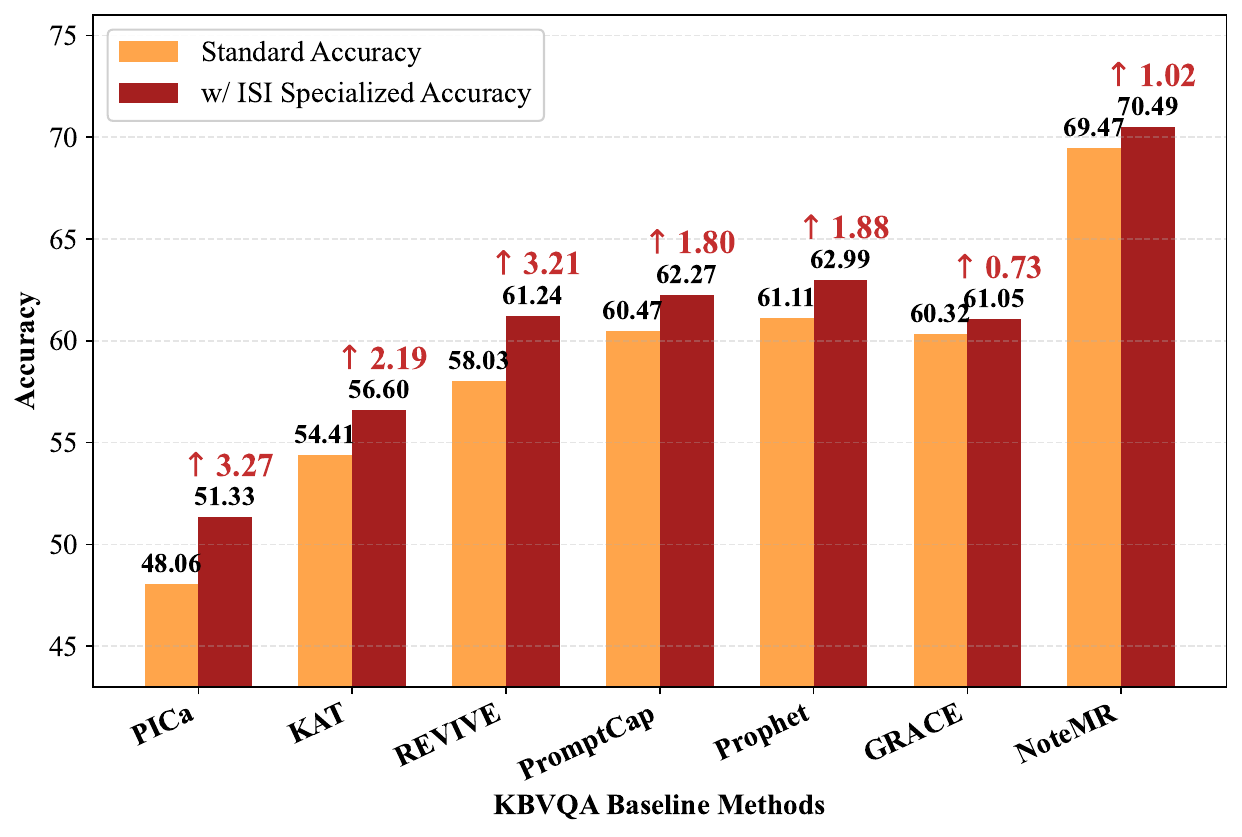}
	\caption{\textbf{Performance comparison of KBVQA methods with the ISI module.} The orange bars and red bars mean the performance under the standard metric and the specialized metric. The numbers with the red arrow indicate the performance gains with the ISI module.}
	\label{fig14}
\end{figure}

\begin{table}[!t]
	\centering
	\setlength{\tabcolsep}{2pt}
	
	\caption{\textbf{Performance Comparison of KBVQA methods with ISI module on \textsc{OKVQA} test set}.Note that a specialized metric is used to evaluate the capacity of ISI module in reliability improvement.}
	\begin{tabular}{l|ccc}
	\toprule[1.1pt]
	Method & Venue & Standard Accuracy &  \begin{tabular}{c}
		\textit{w/} ISI \\  
		Specialized Accuracy  
	\end{tabular} \\
	\midrule
	PICa~\cite{76} & AAAI'22 &48.06 & \cellcolor{gray!20} 51.33  \\
	KAT~\cite{77} & NAACL'22 & 54.41 & \cellcolor{gray!20} 56.60  \\
	REVIVE~\cite{78} & NeurIPS'22 & 58.03 & \cellcolor{gray!20} 61.24  \\
	PromptCap~\cite{79} & ICCV'23 & 60.47 & \cellcolor{gray!20} 62.27  \\
	Prophet~\cite{80} & CVPR'23 & 61.11 & \cellcolor{gray!20} 62.99  \\
    GRACE~\cite{68} & ECCV'24 & 60.32 & \cellcolor{gray!20} 61.05  \\
    NoteMR~\cite{92} & CVPR'25 & 69.47 & \cellcolor{gray!20} 70.49  \\
	\bottomrule[1.1pt]
	\end{tabular}
	\label{tab7}
\end{table}

We also examine the ISI module’s effectiveness during inference by incorporating it into KBVQA models and evaluating on \textsc{OK-VQA}~\citep{74} test set. 
As shown in \cref{tab7} and \cref{fig14}, the ISI module effectively filters unrelated QI inputs that may lead to erroneous answers, improving prediction reliability. 
However, the improvement is limited because the ISI module’s training scale is much smaller than that of LLMs, resulting in weaker multi-modal representation capacity for large-model scenarios.

\section{Conclusion and Discussion}

In this paper, we propose Optimized \underline{Q}uestion-\underline{I}mage \underline{R}elation \underline{L}earning (QIRL), a generation-based self-supervised framework designed to enhance the robustness of VQA. To improve question-image (QI) correlation learning, we introduce a novel generation strategy that automatically produces highly irrelevant QI pairs. Specifically, a Negative Image Generation (\textsc{NIG}) module is devised to generate these irrelevant QI pairs without human supervision. Furthermore, to boost model robustness, we propose an identification module that assesses input relevance and determines whether the model should yield a prediction or abstain during inference. This Irrelevant Sample Identification (\textsc{ISI}) module guides the VQA system in handling irrational inputs that do not correspond to the intended QI pairs via filtering out this kind of input and outputting "abstain". 

Although our method outperforms the baseline and achieves performance comparable to previous approaches, several limitations remain. First, the evaluation protocol for robustness is not yet sufficiently comprehensive. Second, while highly irrelevant QI pairs are successfully generated, further exploration of the bias in QI correlation learning is warranted. Finally, given that large models also suffer from similar bias problems, developing debiasing strategies for VQA methods using large models remains a promising topic for future research.

To address these limitations, we propose several avenues for future research: 
1) developing a more robust and fair evaluation protocol to better validate the proposed method; 
2) extending the debiasing approach from solely addressing language bias to also mitigating vision bias, thereby reducing multi-modal bias; and 
3) incorporating the debiasing concepts into VQA methods that employ large models to further enhance their reliability. 
With these improvements, we anticipate further reductions in language bias and significant gains in model robustness.

\section*{Acknowledge}
This work was supported in part by the Science and Technology Development Fund of Macau, Macau SAR, under Grant Nos. 0035/2023/ITP1 and 0021/2023/RIA1, the National Natural Science Foundation of China, under Grant No. 62271361, and the Hubei Provincial Key Research and Development Program, under Grant No. 2024BAB039.

\balance
\bibliographystyle{ieeetr}
\bibliography{QIRL}



\end{document}